\documentclass{amsart}

\usepackage[utf8]{inputenc}
\usepackage[numbers]{natbib} 
\usepackage{hyperref}
\usepackage{float}
\usepackage{amsmath, amssymb, amsthm}
\usepackage{geometry}
\usepackage{graphicx}
\usepackage{mathrsfs} 
\usepackage{tikz-cd}
\usepackage{mathtools}
\usepackage{amsmath,amssymb,array,booktabs,tabularx}
\usepackage{tikz}
\usepackage{graphicx}
\usetikzlibrary{arrows.meta,positioning}

\theoremstyle{definition}

\theoremstyle{remark}

\theoremstyle{plain}

\theoremstyle{definition}

\title{Geometric state space, memory representation and architectures}
\author{Shang Xu}\author{Pengcheng Wang}

\begin{document}

\begin{abstract}
We introduce a geometric framework for deriving attention architectures. A sequence forms a memory $m\in M_n$, and an attention pairing $\rho_n\colon M_n\to\operatorname{Hom}(A,A)$ determines how that memory acts on each token state. Different representations of this action recover familiar architectures: softmax attention reads individual key--value pairs, linear attention accumulates their contributions in matrix statistics, and Intention fits a regression map from keys to values. The framework also suggests new architectures by changing the geometry of the query--key pairing, the function carried by each value, or the space representing the memory. ImageNet-1K experiments illustrate these constructions through geometric CAFormer variants, a ViT with query-dependent values, and a linear ViT with a wider memory representation.
\end{abstract}
\maketitle
\tableofcontents
\section{Introduction}

A word enters a sentence with a fixed embedding, but the sentence changes how we read it. The word \emph{bank} has the same initial embedding in ``the bank of the river'' and ``the bank approved the loan.'' An attention layer gives it a different state in each context. In a Transformer, this happens through a familiar calculation. A query assigns weights to the keys of the surrounding tokens, and those weights determine how their values are combined \cite{vaswani2017attention}. We take a different starting point for the same calculation. Once the surrounding tokens are fixed, what function do they define on the token being read?

Let $A$ be the space of token states. The vocabulary specifies distinguished points of $A$, and the same space contains the states that tokens acquire in context. Reading a sequence $s=(a_1,\ldots,a_n)$ produces a memory $m_s\in M_n$. We call the map
\[
\rho_n\colon M_n\longrightarrow \operatorname{Hom}(A,A)
\]
the \emph{attention pairing}. A memory determines a transformation of token states. Applying $\rho_n(m_s)$ to each $a_i$ gives a new sequence, to which a pointwise feedforward map can then be applied. The next layer reads these updated states and constructs its own memory. In a causal layer, the state at position $i$ is updated using the memory of the prefix $(a_1,\ldots,a_i)$.

The distinction between a memory and the function it represents matters when we ask what an architecture retains. Two collections of keys and values may differ and still give the same response to every query. We study this possibility through a factorization
\[
M_n\longrightarrow E_n\longrightarrow \operatorname{Hom}(A,A),
\]
where $E_n$ contains the quantities used to evaluate the memory. A representation is faithful on a family of memories if every two distinct memories in that family can be distinguished by some input. This gives a precise form to a practical design question. Which distinctions can a layer preserve at a given cost in storage and evaluation?

Several established architectures give different answers. Softmax attention retains the individual keys and values for the query to read. Linear attention accumulates their contributions in fixed-dimensional matrix and vector statistics \cite{katharopoulos2020transformers}. Intention uses the keys and values to fit a regression function, which it then evaluates at the query \cite{garnelo2023exploring}. Tensor Product Attention introduces structure in the projections themselves through low-rank factorizations \cite{zhang2025tensor}. The comparison also clarifies the relation with recurrent networks, where the accumulated state can be passed directly to a prediction map.

We use these distinctions to design and test attention layers on ImageNet-1K. In CAFormer-S18, we compare ordinary attention with three alternatives: matching lines in projective space, matching four-dimensional subspaces, and allowing two queries to retrieve different values from a shared key. All three variants come within 0.73 percentage points of the baseline at their respective last shared epochs. In ViT-5-Small/16, we let each source token contribute an affine function of the receiving query. This KV-action model reaches 70.96\% top-1 accuracy at epoch~60, compared with 69.50\% for the baseline. The additional projection and a change in score scale are both part of the tested design. Finally, we widen the key space of a linear ViT while keeping its value width fixed. The resulting head has $2.25$ times as many memory coordinates, and the model leads its baseline at all 26 shared epochs despite having slightly fewer learned parameters. These single-seed comparisons support the use of memory geometry as a design principle. The dimension count identifies which distinctions a layer has room to preserve. Which of those distinctions it learns to use remains an empirical question.

\textbf{Acknowledgments.} Thanks Ziang Li for very useful discussions.

\textbf{AI disclosures.} In this first version of paper, ChatGPT 6 Astra is used for literature searches, computation check, language polishing and some experiment coding. The ideas and mathematical interference are original and the author takes full responsibility for its correctness.

\section{Formalization of Characters, Knowledge and Intelligence and Attention}
In this section we explain our abstraction of learning theory to both computer scientists and mathematicians. Subsection \ref{computer scientiest} is for computer scientists to quickly understand our setting, and \ref{mathematical formalizing} is a categorical formalism for mathematicians. 

\subsection{An intuitive formulation}
\label{computer scientiest}

We first describe the framework using vectors and functions. Let $A$
be a space of \emph{word states}; for intuition, one may take
$A\subseteq\mathbb R^d$. A vocabulary $W$ is represented by an embedding
$\iota:W\hookrightarrow A$. The point $\iota(w)$ is the initial state
of the word $w$, before its meaning is influenced by a particular
sentence. Other points of $A$ may represent contextualized word
states and need not correspond to individual vocabulary items.
For example, the initial representation of ``bank'' can be
transformed into different states when it occurs in a sentence about
a river or a financial institution. Both the initial and contextualized
representations belong to the same space $A$.

A sentence of length $N$ is represented by a sequence
$s=(a_1,\ldots,a_N)\in A^N$. Initially, its coordinates are vocabulary
embeddings. After processing, they are updated word states that
reflect the influence of the surrounding words.

For each $n\geq0$, let $M_n$ be a \emph{compressed memory space}.
An element $m_n\in M_n$ represents the information retained after
reading $n$ word states. A memory may be represented by a vector,
a matrix, or another structured object. Its size and structure are
part of the architectural design and may depend on $n$.
Information is accumulated through update maps
\[
\mu_n:M_n\times A\longrightarrow M_{n+1}.
\]
Starting from a fixed initial memory $m_0\in M_0$, reading a sequence
$(a_1,\ldots,a_n)$ gives the recurrence
\[
m_{i+1}=\mu_i(m_i,a_{i+1}),
\qquad 0\leq i<n.
\]
The resulting memory defines a compression map $c_n:A^n\to M_n$,
with $c_n(a_1,\ldots,a_n)=m_n$. The choice of memory space and update
rule determines which features of the sequence are retained.

The next ingredient specifies how memory changes word states.
We introduce a map
\[
\rho_n:M_n\longrightarrow\operatorname{Hom}(A,A),
\]
where $\operatorname{Hom}(A,A)$ denotes the functions from $A$ to
itself in this intuitive formulation. Thus, each memory $m\in M_n$
determines a transformation $\rho_n(m):A\to A$. Applying this
transformation to a word state $a$ produces the updated state
$\rho_n(m)(a)$. These transformations may be nonlinear.

We call $\rho_n$ an \emph{attention pairing}. Equivalently, it can be
written as a function taking the memory and the word state as two
inputs:
\[
\widetilde\rho_n:M_n\times A\longrightarrow A,
\qquad
\widetilde\rho_n(m,a)=\rho_n(m)(a).
\]
For a sentence $s\in A^n$, its action on a word is therefore
$s(a)=\rho_n(c_n(s))(a)$. The sentence first determines a memory,
and that memory determines how the word state changes.

We distinguish a stored memory from the transformation it induces.
The space $M_n$ describes the stored states, whereas
\[
\mathcal R_n
=
\rho_n(M_n)
=
\{\rho_n(m):m\in M_n\}
\subseteq\operatorname{Hom}(A,A)
\]
describes their \emph{memory representation}. Each element of
$\mathcal R_n$ is a function that transforms word states. Two
different memories may have the same representation: if
$\rho_n(m)=\rho_n(m')$, they act identically on every word state,
even though their stored data differ.

The action of a memory may be determined by a representation
with a fixed number of coordinates. To express this, choose a
space $E\subseteq\mathbb R^p$, where $p$ is independent of the
sequence length, together with encoding maps $\lambda_n:M_n\to E$
and a shared decoding map
$\Psi:E\to\operatorname{Hom}(A,A)$. We require the factorization
\[
M_n
\xrightarrow{\lambda_n}
E
\xrightarrow{\Psi}
\operatorname{Hom}(A,A),
\qquad
\rho_n=\Psi\circ\lambda_n.
\]
Thus, the code $e=\lambda_n(m)$ determines the transformation
associated with $m$. The decoder $\Psi$ is a common rule for
turning this code into an action on word states:
\[
\rho_n(m)(a)=\Psi(\lambda_n(m))(a).
\]
In an implementation, this means evaluating a function of the
two inputs $e$ and $a$; it does not require storing a table of
the transformation's values.

We call this a \emph{uniformly finite-dimensional memory
representation}. The same representation space $E$ and decoder
$\Psi$ serve every sequence length, although the encoding map
$\lambda_n$ may depend on $n$. For example, a code may consist
of a matrix and a vector whose dimensions remain fixed as more
words are read. The induced transformation of word states may
still be nonlinear.

For a sequence $s\in A^n$, the representation used to evaluate
words is therefore $e_s=\lambda_n(c_n(s))$, and its action is
$s(a)=\Psi(e_s)(a)$. The sequence determines a memory, the memory
determines a code in $E$, and that code determines how word
states are transformed.

An \emph{attention layer} applies this construction to every
coordinate of the input sequence. Given $s=(a_1,\ldots,a_N)$,
first compute the total memory $m_s=c_N(s)$. Then use the
transformation $\rho_N(m_s)$ to update all word states:
\[
L_N:A^N\longrightarrow A^N,
\qquad
L_N(s)
=
\bigl(
\rho_N(m_s)(a_1),
\ldots,
\rho_N(m_s)(a_N)
\bigr).
\]
Each word contributes to the shared memory, and the shared memory
acts on every word. This provides the mechanism through which
the words influence one another. The output is again a sequence
in $A^N$, so it can be processed by another attention layer.

The construction above uses the memory of the entire sequence.
For \emph{causal attention}, the output at position $i$ must use
only the first $i$ input states. Writing
$s_{\leq i}=(a_1,\ldots,a_i)$, define
\[
a_i'
=
\rho_i\bigl(c_i(s_{\leq i})\bigr)(a_i),
\qquad
C_N(s)=(a_1',\ldots,a_N').
\]
Thus, each position uses the memory of the prefix ending at that
position. In particular, the current word may influence its own
updated state, but later words cannot influence it. The resulting
family of causal layers satisfies
\[
\bigl(C_N(s)\bigr)_{\leq i}=C_i(s_{\leq i}).
\]
Appending more words to the input therefore leaves the previously
determined output prefix unchanged. Computing a memory sequentially
does not by itself establish this property: causality also requires
each output to use only its corresponding prefix memory.

After attention updates the word states, a \emph{feedforward function}
$f:A\to A$ is applied independently to each position. The same
function is used for every word. If
$L_N(s)=(a_1',\ldots,a_N')$, the resulting block is
\[
T_N(s)
=
\bigl(f(a_1'),\ldots,f(a_N')\bigr),
\qquad
T_N=f^{\times N}\circ L_N.
\]
For causal attention, the block is instead
$T_N=f^{\times N}\circ C_N$.
Applying $f$ independently at each position preserves causality. Repeated application of a fixed block gives a
\emph{stack of attention layers}. Starting from $s^{(0)}=s$, define
\[
s^{(k+1)}=T_N(s^{(k)}),
\qquad
s^{(k)}=T_N^k(s).
\]
Here $N$ is the sequence length, while $k$ counts the refinement
steps. Each step accumulates memory from the current word states,
uses attention to transform those states, and then applies the
feedforward function. The stopping and convergence conditions
below concern this complete block operator $T_N$.

What could be mathematically interesting is the convergence of this stack. A natural stopping condition is that another application of the
layer no longer changes the sequence. Such a state $s_*$ satisfies
$T_N(s_*)=s_*$ and is called a \emph{fixed point}. Exact stabilization
at step $k$ means $s^{(k+1)}=s^{(k)}$. More generally, we seek
operators whose repeated applications converge:
\[
T_N^k(s)\longrightarrow s_\infty,
\qquad
T_N(s_\infty)=s_\infty.
\]
When $A\subseteq\mathbb R^d$, convergence can be understood using
the usual topology on sequences of vectors. If $T_N$ is continuous,
any limit of its iterates is a fixed point. Convergence itself is
an additional requirement on the chosen memory updates and
attention pairings; it does not follow from their definitions.
Different input sequences may converge to different stable states.
In a numerical implementation, one may use a stopping rule such as
$\|T_N(s^{(k)})-s^{(k)}\|\leq\varepsilon$, with a chosen norm and
tolerance $\varepsilon>0$.

Finally, positional information can be incorporated before the
first attention layer. Choose a map $\phi_i:A\to A$ for each
position $i\geq1$, and define the coordinatewise transformation
\[
\Phi_N:A^N\longrightarrow A^N,
\qquad
\Phi_N(a_1,\ldots,a_N)
=
\bigl(\phi_1(a_1),\ldots,\phi_N(a_N)\bigr).
\]
The same vocabulary word can then have different initial states
at different positions. For example, when $A=\mathbb R^d$, additive positional
encoding is the choice $\phi_i(a)=a+p_i$, where $p_i$ is a positional
vector. With positional encoding, the initial sequence becomes
$s^{(0)}=\Phi_N(s)$, and the stack produces $T_N^k(\Phi_N(s))$.
Using the same map $\phi_i$ for position $i$ across all sequence
lengths makes this initialization compatible with prefixes and
therefore preserves causality when followed by causal attention
layers.

\subsection{A categorical formalization}
\label{mathematical formalizing}
This subsection is for mathematics
For mathematicians, the following categorical formulation may be preferable, as it allows the spaces above to be replaced by more general geometric or algebraic objects.

Let $\mathcal C$ be a locally small category with finite products and
terminal object $1$. Fix an object $A\in\mathcal C$, called the
\emph{word space}. A vocabulary is specified by a set $W$ together with
an injection
$\iota:W\hookrightarrow\operatorname{Hom}_{\mathcal C}(1,A)$.
The distinguished states $\iota(w)$ represent raw words, while other
states of $A$ may represent words under contextual influence. For
$n\geq0$, the object of sequences of length $n$ is $A^n$, with $A^0=1$.
The vocabulary determines raw sequences among the global elements
of $A^n$.

A \emph{compressed memory system} consists of a sequence of objects
$(M_n)_{n\geq0}$, an initial memory $m_0:1\to M_0$, and update morphisms
\[
\mu_n:M_n\times A\longrightarrow M_{n+1}.
\]
These updates induce compression morphisms $c_n:A^n\to M_n$, defined
recursively by
\[
c_0=m_0,
\qquad
c_{n+1}=\mu_n\circ(c_n\times\operatorname{id}_A),
\]
under the canonical identification $A^{n+1}\cong A^n\times A$. $c_n$ assigns to a sequence its total compressed memory.

A memory determines a transformation of word states. Whenever the
internal Hom object
$\underline{\operatorname{Hom}}_{\mathcal C}(A,A)$ exists, this
assignment is expressed by a morphism
\[
\rho_n:M_n\longrightarrow
\underline{\operatorname{Hom}}_{\mathcal C}(A,A).
\]
For a memory state $m$, the corresponding transformation sends a word
state $a$ to $\rho_n(m)(a)$. We call this assignment an
\emph{attention pairing}.

To formulate the same notion without assuming the existence of an
internal Hom object, write
$h_X=\operatorname{Hom}_{\mathcal C}(-,X)$ and define the presheaf
$\mathscr E_A:\mathcal C^{\mathrm{op}}\to\mathbf{Set}$ by
\[
\mathscr E_A(X)=\operatorname{Hom}_{\mathcal C}(X\times A,A),
\qquad
\mathscr E_A(f)(g)=g\circ(f\times\operatorname{id}_A).
\]
In general, an attention pairing is a natural transformation
$\rho_n:h_{M_n}\Rightarrow\mathscr E_A$. When $\mathscr E_A$ is
representable, this recovers the internal-Hom formulation above.
By the Yoneda lemma, such a natural transformation is equivalently
specified by a morphism
\[
\widetilde\rho_n:M_n\times A\longrightarrow A,
\qquad
\widetilde\rho_n
=(\rho_n)_{M_n}(\operatorname{id}_{M_n}).
\]
Conversely, for every $m:X\to M_n$, its curried form is given by
$(\rho_n)_X(m)=\widetilde\rho_n\circ(m\times\operatorname{id}_A)$.

We distinguish the memory space from its representation by
transformations of the word space. The \emph{memory representation}
associated with $\rho_n$ is the image presheaf
$\mathscr R_n\subseteq\mathscr E_A$, defined by
\[
\mathscr R_n(X)
=
\operatorname{im}\bigl((\rho_n)_X\bigr)
=
\left\{
\widetilde\rho_n\circ(m\times\operatorname{id}_A)
:
m\in\operatorname{Hom}_{\mathcal C}(X,M_n)
\right\}.
\]
Naturality of $\rho_n$ makes $\mathscr R_n$ a subfunctor of
$\mathscr E_A$. In particular,
$\mathscr R_n(1)\subseteq\operatorname{Hom}_{\mathcal C}(A,A)$
is the set of word-state transformations induced by global memory
states. Distinct memory states may represent the same transformation.
The image is taken in the presheaf category, without requiring
image objects in $\mathcal C$.

A \emph{common representation space} for the attention pairings
consists of an object $E\in\mathcal C$, morphisms
$\lambda_n:M_n\to E$, and a natural transformation
$\Psi:h_E\Rightarrow\mathscr E_A$, independent of $n$, such that
\[
\rho_n=\Psi\circ h_{\lambda_n}.
\]
Here $h_{\lambda_n}:h_{M_n}\Rightarrow h_E$ is induced by
postcomposition with $\lambda_n$. Whenever the internal Hom
object exists, this is the factorization
\[
M_n
\xrightarrow{\lambda_n}
E
\xrightarrow{\Psi}
\underline{\operatorname{Hom}}_{\mathcal C}(A,A).
\]
Thus, $\lambda_n$ represents a memory in the common space $E$,
and $\Psi$ assigns to that representation its action on word states.

Equivalently, if $\widetilde\Psi:E\times A\to A$ is the
uncurried form of $\Psi$, the factorization condition is
\[
\widetilde\rho_n
=
\widetilde\Psi\circ
(\lambda_n\times\operatorname{id}_A).
\]
The representation maps need not distinguish all memory states.

In geometric realizations where dimension is defined, we call
such a factorization a \emph{uniformly finite-dimensional memory
representation} when $E$ has finite dimension, independent of
the sequence length $n$. This bounds the size of the intermediate
representation used to determine the word-state transformation.
When $A$ is a vector space, the induced transformations may still
be nonlinear, even if the elements of $E$ are represented by
linear operators.

For a sequence $s:1\to A^n$, let $m_s=c_n\circ s$ be its compressed
memory. Its action on a word state $a:1\to A$ is
$s(a):=\widetilde\rho_n\circ\langle m_s,a\rangle$.
Thus, the action of the sequence on a word factors through the
compressed memory of that sequence.

For $N\geq1$, let $\pi_i^N:A^N\to A$ denote the $i$-th coordinate
projection. The associated \emph{full-context attention layer} is
\[
L_N
=
\left\langle
\widetilde\rho_N\circ\langle c_N,\pi_i^N\rangle
\right\rangle_{i=1}^{N}
:
A^N\longrightarrow A^N.
\]
Equivalently, for $s=(a_1,\ldots,a_N)$,
\[
L_N(s)
=
\bigl(
\widetilde\rho_N(c_N(s),a_1),
\ldots,
\widetilde\rho_N(c_N(s),a_N)
\bigr).
\]
Every word contributes to the shared memory, and that memory acts
on every coordinate of the sequence.

To introduce \emph{causality}, let $p_i^N:A^N\to A^i$ be the projection
onto the first $i$ coordinates. A family of morphisms
$T_N:A^N\to A^N$ is called causal if
\[
p_i^N\circ T_N=T_i\circ p_i^N
\qquad
(1\leq i\leq N).
\]
This condition requires the transformed prefix to depend only on
the input prefix, consistently across sequence lengths. The
\emph{causal attention layer} associated with the memory system is
\[
C_N
=
\left\langle
\widetilde\rho_i\circ
\left\langle c_i\circ p_i^N,\pi_i^N\right\rangle
\right\rangle_{i=1}^{N}
:
A^N\longrightarrow A^N.
\]
Its $i$-th output is
$\widetilde\rho_i(c_i(a_1,\ldots,a_i),a_i)$.
In particular, each word is transformed using the memory of the
prefix ending at that word. The family $(C_N)_{N\geq1}$ satisfies
$p_i^N\circ C_N=C_i\circ p_i^N$.

A \emph{feedforward transformation} is a morphism $f:A\to A$.
Its coordinatewise extension is
\[
f^{\times N}:A^N\longrightarrow A^N,
\qquad
(a_1,\ldots,a_N)\longmapsto
\bigl(f(a_1),\ldots,f(a_N)\bigr).
\]
Composing this transformation with attention gives the block operator
\[
T_N=f^{\times N}\circ L_N
\]
for full-context attention, or $T_N=f^{\times N}\circ C_N$ for
causal attention. Since
$p_i^N\circ f^{\times N}=f^{\times i}\circ p_i^N$,
the causal block satisfies $p_i^N\circ T_N=T_i\circ p_i^N$.
Repeated application of a fixed block defines the
\emph{iterated attention stack}:
$T_N^0=\operatorname{id}_{A^N}$ and
$T_N^{k+1}=T_N\circ T_N^k$.
Starting from $s$, the state at depth $k$ is $s^{(k)}=T_N^k(s)$.
The stabilization and convergence conditions below apply to this
composite operator. In the causal case,
\[
p_i^N\circ T_N^k=T_i^k\circ p_i^N
\qquad
(k\geq0,\ 1\leq i\leq N).
\]

A global state $s_*:1\to A^N$ is \emph{stable} if
$T_N\circ s_*=s_*$. Accordingly, exact stabilization of an initial
state $s$ at depth $k$ means
$T_N^{k+1}\circ s=T_N^k\circ s$.
When the equalizer exists, the object
$F_N=\operatorname{Eq}(T_N,\operatorname{id}_{A^N})$
represents stable states.

Asymptotic stabilization requires additional convergence data.
Let $S_N=\operatorname{Hom}_{\mathcal C}(1,A^N)$, and equip each
$S_N$ with a Hausdorff topology. Require the induced layer maps
$t_N:S_N\to S_N$, defined by $t_N(s)=T_N\circ s$, and the induced
prefix maps $q_i^N:S_N\to S_i$, defined by
$q_i^N(s)=p_i^N\circ s$, to be continuous. These topologies are
additional structure on the sets of states; the categorical
definitions above do not require a topology on $A$.

We impose the convergence condition that every orbit has a limit:
\[
\forall N\geq1,\ \forall s\in S_N,
\qquad
P_N(s):=\lim_{k\to\infty}t_N^k(s)
\quad\text{exists in }S_N.
\]
This defines a set map $P_N:S_N\to S_N$. Continuity of $t_N$ and
uniqueness of limits imply
\[
t_N\circ P_N=P_N\circ t_N=P_N,
\qquad
P_N^2=P_N,
\qquad
\operatorname{im}P_N=\operatorname{Fix}(t_N).
\]
The limiting state is therefore stable and may depend on the
initial sequence. Continuity of $P_N$ is not part of this condition.
For a causal family, continuity of the prefix maps additionally
gives $q_i^N\circ P_N=P_i\circ q_i^N$.

Positional information can be incorporated by transforming each
word according to its position before applying the attention
layers. Choose a family of endomorphisms
$(\phi_i:A\to A)_{i\geq1}$. For sequences of length $N$, these
define the coordinatewise, or diagonal, transformation
\[
\Phi_N=\phi_1\times\cdots\times\phi_N:A^N\longrightarrow A^N,
\qquad
\Phi_N(a_1,\ldots,a_N)
=
\bigl(\phi_1(a_1),\ldots,\phi_N(a_N)\bigr).
\]
The same word can thus acquire different initial states at
different positions. With positional encoding, the attention
stack acts on $\Phi_N(s)$ and produces $T_N^k(\Phi_N(s))$.
Because the maps $\phi_i$ are chosen consistently across sequence
lengths, one has
$p_j^N\circ\Phi_N=\Phi_j\circ p_j^N$.
Consequently, applying this positional transformation before a
causal attention stack preserves causality.

\section{From the formalization to architectures}

In this section, we specify how the geometric constructions of the knowledge world and intelligence in last section's examples lead to the actual architectures in industry. Our examples cover attention, casual attention, linear attention, tensor attention and vanilla/multiplicative RNN, showing the strong potential ability of our formalization in architectural designs.

\subsection{Feature memory and Softmax attention}
\label{attention}

We realize scaled dot-product attention by specifying a word
space, a key--value memory, and its representation as a family
of transformations of word states. We describe a single attention
head with an output projection, followed by a feedforward map.

Let $A=\mathbb R^d$. Fix a query--key dimension $r$, a value
dimension $v$, and trainable matrices
\[
W_Q,W_K\in\mathbb R^{r\times d},
\qquad
W_V\in\mathbb R^{v\times d},
\qquad
W_O\in\mathbb R^{d\times v}.
\]
For a word state $a\in A$, its query, key, and value are
$W_Qa$, $W_Ka$, and $W_Va$, respectively. The output projection
$W_O$ maps the retrieved value representation back into $A$.

For each $n\geq0$, define
\[
M_n=\mathbb R^{r\times n}\times\mathbb R^{v\times n}.
\]
A memory $(K,V)\in M_n$ stores $n$ keys and their corresponding
values as columns. The initial memory $m_0\in M_0$ is the unique
pair of empty matrices, and the updates are
\[
\mu_n:M_n\times A\longrightarrow M_{n+1},
\qquad
\mu_n((K,V),a)
=
\bigl([K\mid W_Ka],\,[V\mid W_Va]\bigr).
\]
The resulting compression map sends $s=(a_1,\ldots,a_n)$ to
$c_n(s)=(K_n,V_n)$, where
\[
K_n=[W_Ka_1\mid\cdots\mid W_Ka_n],
\qquad
V_n=[W_Va_1\mid\cdots\mid W_Va_n].
\]

\subsubsection{The attention pairing and its memory representation}

We first introduce the attention pairing as a map
\[
\rho_n:M_n\longrightarrow\operatorname{Hom}(A,A),
\]
where, for this example, we take
$\operatorname{Hom}(A,A)=C(A,A)$.
For $n\geq1$, define
$\operatorname{sm}_n:\mathbb R^n\to\mathbb R^n$ by
\[
[\operatorname{sm}_n(z)]_j
=
\frac{\exp(z_j)}{\sum_{\ell=1}^n\exp(z_\ell)}.
\]
The transformation determined by a memory $(K,V)$ is
\[
\rho_n(K,V)(a)
=
W_OV\,
\operatorname{sm}_n
\left(\frac{K^{\mathsf T}W_Qa}{\sqrt r}\right).
\]
At the empty memory, set $\rho_0(m_0)(a)=0$.
The uncurried attention pairing is consequently
$\widetilde\rho_n((K,V),a)=\rho_n(K,V)(a)$.

The memory determines how each word is transformed: its query
is compared with the stored keys, the resulting scores are
normalized by softmax, and the stored values are aggregated
using these weights. Thus, the attention pairing includes
compatibility evaluation, normalization, and value aggregation.
Its output is already a word state in $A$.

For each fixed memory, this transformation is smooth in $a$.
The associated memory representation therefore satisfies
\[
\mathcal R_n
=
\rho_n(M_n)
\subseteq C^\infty(A,A)
\subseteq C(A,A).
\]
Here $M_n$ records the stored key--value arrays, whereas
$\mathcal R_n$ records the transformations those arrays induce. In this realization, the stored pair $(K,V)$ itself supplies a
finite-dimensional code for the represented transformation.
In the factorization notation, one may take $E_n=M_n$,
$\lambda_n=\operatorname{id}_{M_n}$, and $\Psi_n=\rho_n$.
This carrier uses $n(r+v)$ real coordinates. It is therefore
finite-dimensional for each sequence length, but its size grows
with $n$; this storage scheme dimension increase as the sequence length increase.

For a sequence $s=(a_1,\ldots,a_N)$, its total memory
$m_s=c_N(s)$ determines the attention operator
\[
L_N:A^N\longrightarrow A^N,
\qquad
L_N(s)
=
\bigl(
\rho_N(m_s)(a_1),\ldots,\rho_N(m_s)(a_N)
\bigr).
\]
Writing $L_N(s)=(y_1,\ldots,y_N)$ gives
\[
y_i
=
W_O\sum_{j=1}^N\alpha_{ij}W_Va_j,
\qquad
\alpha_{ij}
=
\frac{
\exp\!\left((W_Qa_i)^{\mathsf T}(W_Ka_j)/\sqrt r\right)
}{
\displaystyle\sum_{\ell=1}^N
\exp\!\left((W_Qa_i)^{\mathsf T}(W_Ka_\ell)/\sqrt r\right)
}.
\]
This recovers scaled dot-product self-attention
\cite[Section~3.2.1]{vaswani2017attention}, followed by the output
projection $W_O$.

A feedforward transformation $f:A\to A$ is then applied
independently at each position. For example, choose a hidden
dimension $h$ and define
\[
f(a)=W_2\sigma(W_1a+b_1)+b_2,
\]
where $W_1\in\mathbb R^{h\times d}$,
$W_2\in\mathbb R^{d\times h}$,
$b_1\in\mathbb R^h$, $b_2\in\mathbb R^d$, and $\sigma$ is a
continuous activation applied coordinatewise. The same function
$f$ is used at every position. The attention and feedforward
composition is
\[
T_N=f^{\times N}\circ L_N,
\qquad
T_N(s)=\bigl(f(y_1),\ldots,f(y_N)\bigr).
\]
The smoothness statement for $\mathcal R_n$ concerns the
attention pairing before this additional composition.

These descriptions need not distinguish the same states.
For example, if $P$ is an $n\times n$ permutation matrix, then
\[
\rho_n(KP,VP)=\rho_n(K,V).
\]
Simultaneously permuting the stored keys and their associated
values leaves the represented transformation unchanged. Positional information is incorporated by applying
$\Phi_N=\phi_1\times\cdots\times\phi_N$ to the initial sequence,
where $\phi_i:A\to A$ is the transformation associated with
position $i$. For additive positional embeddings, take
$\phi_i(a)=a+p_i$ with $p_i\in\mathbb R^d$.
The first block acts on $\Phi_N(s)$, so its keys, queries,
and values are computed from these position-dependent word
states. Choosing the same $\phi_i$ for position $i$ across
sequence lengths ensures
$p_j^N\circ\Phi_N=\Phi_j\circ p_j^N$ and preserves causality
when this initialization is followed by causal attention.

\subsubsection{Causal attention}
\label{casual attention}

For causal attention, the word at position $i$ is transformed
using only the memory of the prefix
$s_{\leq i}=(a_1,\ldots,a_i)$. This memory is
$m_i=c_i(s_{\leq i})=(K_i,V_i)$, where $K_i$ and $V_i$ contain
only the first $i$ keys and values. Define
\[
C_N(s)_i
=
\rho_i(m_i)(a_i)
=
W_OV_i\,
\operatorname{sm}_i
\left(\frac{K_i^{\mathsf T}W_Qa_i}{\sqrt r}\right).
\]
The same family of memory representations is used, but position
$i$ accesses the transformation represented by its prefix memory.

Explicitly,
\[
C_N(s)_i
=
W_O\sum_{j=1}^i\alpha_{ij}^{\mathrm{causal}}W_Va_j,
\]
where, for $j\leq i$,
\[
\alpha_{ij}^{\mathrm{causal}}
=
\frac{
\exp\!\left((W_Qa_i)^{\mathsf T}(W_Ka_j)/\sqrt r\right)
}{
\displaystyle\sum_{\ell=1}^i
\exp\!\left((W_Qa_i)^{\mathsf T}(W_Ka_\ell)/\sqrt r\right)
}.
\]
Both normalization and value aggregation involve only the
available prefix. This is the masked scaled dot-product
self-attention computation used in the Transformer decoder
\cite[Section~3.2.3]{vaswani2017attention}.

Let $p_i^N:A^N\to A^i$ be the prefix projection. The family
$(C_N)_{N\geq1}$ satisfies
\[
p_i^N\circ C_N=C_i\circ p_i^N.
\]
The corresponding attention and feedforward block is
$T_N^{\mathrm{causal}}=f^{\times N}\circ C_N$.
Since the feedforward map acts independently and identically
at every position, it preserves prefix compatibility:
\[
p_i^N\circ T_N^{\mathrm{causal}}
=
T_i^{\mathrm{causal}}\circ p_i^N.
\]

\subsection{Linear attention and fixed-dimensional memory representations}
\label{linear attention}

We next realize positive-feature linear attention within our
framework. Each word contributes a key--value pair to memory and
a rank-one matrix to its representation. The accumulated memory
admits a representation in a fixed-dimensional space of linear
operators, from which a normalized transformation of word states
is obtained.

Let $A=\mathbb R^d$ be the word space. Fix a feature dimension $r$,
a value dimension $m$, and matrices
\[
W_Q,W_K\in\mathbb R^{r\times d},
\qquad
W_V\in\mathbb R^{m\times d},
\qquad
W_O\in\mathbb R^{d\times m}.
\]
Choose a continuous feature map
$\varphi:\mathbb R^r\to\mathbb R_{>0}^r$, and define
\[
q(a)=\varphi(W_Qa),
\qquad
k(a)=\varphi(W_Ka),
\qquad
v(a)=W_Va.
\]
Thus, $q(a)$ and $k(a)$ are strictly positive query and key
features, while $v(a)$ is the value associated with $a$.
The output projection $W_O$ returns a retrieved value vector
to the word space $A$.

\subsubsection{Memory spaces and rank-filtered representations}

Define $M_n=(\mathbb R_{>0}^r\times\mathbb R^m)^n$, with
$M_0=\{\ast\}$. A memory is an ordered sequence
$x=((k_1,v_1),\ldots,(k_n,v_n))$. The update
$\mu_n:M_n\times A\to M_{n+1}$ appends $(k(a),v(a))$ to $x$,
so $c_n(a_1,\ldots,a_n)=((k(a_j),v(a_j)))_{j=1}^n$.

The contribution of a word state $a\in A$ to the matrix
representation is
\[
g(a)=\bigl[k(a)\mid k(a)v(a)^{\mathsf T}\bigr]
\in\mathbb R^{r\times(m+1)}.
\]
Since $g(a)=k(a)[1\mid v(a)^{\mathsf T}]$, this matrix has rank
one. The word state belongs to $A$, while $g(a)$ describes its
contribution to the memory representation.

For $n\geq0$, define
\[
E_n
=
\{0\}\cup
\left\{
H=[z\mid S]\in\mathbb R^{r\times(m+1)}
:
z\in\mathbb R_{>0}^r,\ 
\operatorname{rank}H\leq n
\right\}.
\]
In particular, $E_0=\{0\}$ and $E_n\subseteq E_{n+1}$.
The encoding $\lambda_n:M_n\to E_n$ is
$\lambda_n(x)=\sum_{j=1}^n[k_j\mid k_jv_j^{\mathsf T}]$,
with $\lambda_0(\ast)=0$. The updates of the representation are
\[
\bar\mu_n:E_n\times A\longrightarrow E_{n+1},
\qquad
\bar\mu_n(H,a)=H+g(a).
\]
They satisfy
$\lambda_{n+1}(\mu_n(x,a))=\bar\mu_n(\lambda_n(x),a)$.
These maps are well-defined because the first column remains
strictly positive after a nonempty update and
\[
\operatorname{rank}(H+g(a))
\leq\operatorname{rank}H+1
\leq n+1.
\]

Starting from $H_0=0$, a sequence $s=(a_1,\ldots,a_n)$ produces
the representation $H_n=\lambda_n(c_n(s))=[z_n\mid S_n]$, where
\[
z_n=\sum_{j=1}^n k(a_j),
\qquad
S_n=\sum_{j=1}^n k(a_j)v(a_j)^{\mathsf T}.
\]
Equivalently, the updates are
$z_{n+1}=z_n+k(a_{n+1})$ and
$S_{n+1}=S_n+k(a_{n+1})v(a_{n+1})^{\mathsf T}$.
The rank bound records the number of rank-one contributions,
while every memory representation has the same ambient
matrix shape.

\subsubsection{Memory representation and attention pairing}

Take $\operatorname{Hom}(A,A)=C^\infty(A,A)$. We define the attention
pairing
\[
\rho_n:M_n\longrightarrow\operatorname{Hom}(A,A)
\]
through a common finite-dimensional representation space.

Let $Q=\mathbb R^r$. For $H=[z\mid S]$, associate the linear map
\[
\ell_H:Q\longrightarrow\mathbb R^m\oplus\mathbb R,
\qquad
\ell_H(\xi)=\bigl(S^{\mathsf T}\xi,\ z^{\mathsf T}\xi\bigr).
\]
Define the common representation space
\[
E=
\left\{
\ell_{[z\mid S]}:
S\in\mathbb R^{r\times m},\
z\in\mathbb R_{\geq0}^r
\right\}
\subseteq
\operatorname{Hom}_{\mathbb R}
\bigl(Q,\mathbb R^m\oplus\mathbb R\bigr).
\]
This is a subset of a linear-operator space of dimension
$r(m+1)$, independent of the sequence length. The representation
maps from the matrix spaces are $\eta_n:E_n\to E$, with
$\eta_n(H)=\ell_H$. Thus a memory $x\in M_n$ is represented
in $E$ by $\eta_n(\lambda_n(x))$.

Fix $\varepsilon>0$ and define
\[
\nu_\varepsilon:
\mathbb R^m\times\mathbb R_{\geq0}\longrightarrow\mathbb R^m,
\qquad
\nu_\varepsilon(u,c)=\frac{u}{\varepsilon+c}.
\]
The decoding map $\Psi:E\to\operatorname{Hom}(A,A)$ is
\[
\Psi(\ell)(a)
=
W_O\,\nu_\varepsilon\bigl(\ell(q(a))\bigr).
\]
This expression is well-defined because the last coordinate
of $\ell(q(a))$ is nonnegative. The attention pairing therefore
factors as
\[
M_n
\xrightarrow{\lambda_n}
E_n
\xrightarrow{\eta_n}
E
\xrightarrow{\Psi}
\operatorname{Hom}(A,A),
\qquad
\rho_n=\Psi\circ\eta_n\circ\lambda_n.
\]
Explicitly, for $\lambda_n(x)=[z\mid S]$,
\[
\rho_n(x)(a)
=
W_O\frac{S^{\mathsf T}q(a)}
{\varepsilon+z^{\mathsf T}q(a)}.
\]
Its uncurried form is
$\widetilde\rho_n(x,a)=\rho_n(x)(a)$.
The positive constant $\varepsilon$ also defines the readout
continuously at the zero representation, where its value is zero.

The associated memory representation is
$\mathcal R_n=\Psi(\eta_n(\lambda_n(M_n)))\subseteq C^\infty(A,A)$.
Although the intermediate representations $\ell_H$ are linear
maps on the query-feature space $Q$, the induced transformations
of word states are generally nonlinear because they include
feature extraction and normalization. The space $E$ and the
decoder $\Psi$ are the same for every $n$, giving a uniformly
finite-dimensional memory representation.

Moreover, this representation supports direct incremental
updates:
\[
\eta_{n+1}(\bar\mu_n(H,a))
=
\eta_n(H)+\ell_{g(a)}.
\]
Thus, the represented memory can be updated by adding the
linear operator associated with the new word contribution.
Distinct ordered memories may have the same matrix representation
and hence the same word transformation; keeping $M_n$ separate
from $E_n$ makes this possible failure of faithfulness explicit.

\subsubsection{Attention and feedforward composition}

For a sequence $s=(a_1,\ldots,a_N)$, let
$x_N=c_N(s)$ and $H_N=\lambda_N(x_N)=[z_N\mid S_N]$.
Applying its represented transformation to each word gives
\[
L_N(s)
=
\bigl(
\rho_N(x_N)(a_1),\ldots,\rho_N(x_N)(a_N)
\bigr).
\]
Writing $L_N(s)=(y_1,\ldots,y_N)$, we obtain
\[
y_i
=
W_O\frac{S_N^{\mathsf T}q(a_i)}
{\varepsilon+z_N^{\mathsf T}q(a_i)}
=
W_O
\frac{
\displaystyle\sum_{j=1}^N
\bigl(q(a_i)^{\mathsf T}k(a_j)\bigr)v(a_j)
}{
\displaystyle\varepsilon+
\sum_{j=1}^N q(a_i)^{\mathsf T}k(a_j)
}.
\]
The weights are determined by dot products of positive query
and key features. For a nonzero memory, setting $\varepsilon=0$
gives positive weights summing to one and recovers the normalized
positive-feature linear attention computation of
\cite[Section~3.2]{katharopoulos2020transformers}, followed by
the output projection $W_O$. With $\varepsilon>0$, the formula
is a regularized version whose weights sum to less than one.

A feedforward map $f:A\to A$, applied independently and identically
at every position, gives the block
\[
T_N=f^{\times N}\circ L_N,
\qquad
T_N(s)=\bigl(f(y_1),\ldots,f(y_N)\bigr).
\]

The computational advantage follows from reusing $S_N$ and
$z_N$ for every query. After computing the features and values,
each representation update and each query readout requires
$O(rm)$ operations, before the output projection and feedforward
map. Thus, processing $N$ words requires $O(Nrm)$ operations
for these attention computations, while the running attention
representation occupies $O(rm)$ storage. Here ``linear'' refers
to the dependence on sequence length.

\subsubsection{Causal linear attention}

For causal attention, position $i$ uses the prefix memory
$x_i=c_i(a_1,\ldots,a_i)$, represented by
$H_i=\lambda_i(x_i)=[z_i\mid S_i]$. These representations are
computed recursively by
\[
z_i=z_{i-1}+k(a_i),
\qquad
S_i=S_{i-1}+k(a_i)v(a_i)^{\mathsf T},
\qquad
z_0=0,\quad S_0=0.
\]
The causal attention operator is
\[
C_N(s)_i
=
\rho_i(x_i)(a_i)
=
W_O\frac{S_i^{\mathsf T}q(a_i)}
{\varepsilon+z_i^{\mathsf T}q(a_i)}.
\]
Equivalently,
\[
C_N(s)_i
=
W_O
\frac{
\displaystyle\sum_{j=1}^i
\bigl(q(a_i)^{\mathsf T}k(a_j)\bigr)v(a_j)
}{
\displaystyle\varepsilon+
\sum_{j=1}^i q(a_i)^{\mathsf T}k(a_j)
}.
\]
Both the numerator and the denominator use only the available
prefix. At $\varepsilon=0$, this is the causal linear attention
readout of \cite[Section~3.3]{katharopoulos2020transformers}.

Writing $p_i^N:A^N\to A^i$ for the prefix projection, we have
$p_i^N\circ C_N=C_i\circ p_i^N$. The corresponding feedforward
composition
$T_N^{\mathrm{causal}}=f^{\times N}\circ C_N$
also satisfies
\[
p_i^N\circ T_N^{\mathrm{causal}}
=
T_i^{\mathrm{causal}}\circ p_i^N.
\]
Although the accumulated representation is a sum of contributions,
the memory available at each position is determined by its
particular prefix.

For positional transformations acting on word states, one may
apply $\Phi_N=\phi_1\times\cdots\times\phi_N$ before the attention
computation at a chosen layer. The query, key, and value maps
are then evaluated on $\phi_i(a_i)$ at position $i$.
This leaves the matrix shape and the common representation
space $E$ unchanged. When the positional maps are compatible
with prefix projections, their composition with the causal
block preserves causality.

\subsection{Tensor product attention}
\label{tensor attention}

Tensor Product Attention represents the queries, keys, and values
of each word by low-rank matrices, with one row for each attention
head \cite{zhang2025tensor}. In our framework, the word space remains
$A=\mathbb R^d$. The memory stores the factors of the key and value
matrices, and the attention pairing turns this memory into a
transformation of word states.

\subsubsection{Tensor encodings and memory}

Fix $h$ attention heads, each with feature dimension $r$.
A word state $a\in A$ determines three matrices
$Q(a),K(a),V(a)\in\mathbb R^{h\times r}$.
Their $\ell$-th rows are the query, key, and value vectors used
by head $\ell$.

Choose ranks $R_Q,R_K,R_V$. For each $X\in\{Q,K,V\}$, let
$e_X$ be a learned factor encoder with
\[
e_X(a)=\bigl(U_X(a),B_X(a)\bigr),
\qquad
U_X(a)\in\mathbb R^{h\times R_X},
\quad
B_X(a)\in\mathbb R^{r\times R_X}.
\]
The corresponding tensor is
\[
X(a)=\frac1{R_X}U_X(a)B_X(a)^{\mathsf T}.
\]
Hence $\operatorname{rank}X(a)\leq R_X$.
The factor maps may, for example, be linear maps whose outputs
are reshaped into the indicated matrices. We assume they are
smooth.

Write
$\mathcal F_X=\mathbb R^{h\times R_X}\times
\mathbb R^{r\times R_X}$
for the space of factors of type $X$. The memory spaces are
\[
M_n=(\mathcal F_K\times\mathcal F_V)^n,
\qquad M_0=\{*\}.
\]
A memory therefore contains the key and value factors of the
words already read. Query factors are computed from the word
being evaluated and need not be retained in the memory.

The update appends the factors of the new word:
\[
\mu_n((m_1,\ldots,m_n),a)
=
\bigl(m_1,\ldots,m_n,(e_K(a),e_V(a))\bigr).
\]
Starting from the empty memory, a sequence
$s=(a_1,\ldots,a_n)$ gives
\[
c_n(s)
=
\bigl((e_K(a_1),e_V(a_1)),\ldots,
      (e_K(a_n),e_V(a_n))\bigr).
\]

This cache contains $n(R_K+R_V)(h+r)$ scalar entries.
Storing the reconstructed key and value matrices directly would
require $2nhr$ entries. Thus, suitable rank choices reduce the
storage required per word, while the total memory still grows
with the number of words.

\subsubsection{The transformation represented by a memory}

We introduce the attention pairing as
\[
\rho_n:M_n\longrightarrow\operatorname{Hom}(A,A),
\]
where $\operatorname{Hom}(A,A)=C^\infty(A,A)$.
For a stored memory $m$, denote its reconstructed key and value
matrices by $K_1,V_1,\ldots,K_n,V_n$. These matrices specify the
readout below, but need not be explicitly materialized in an
implementation.

Let $q_\ell(a)\in\mathbb R^r$ be the $\ell$-th row of $Q(a)$,
regarded as a column vector. Similarly, let
$k_{j,\ell},v_{j,\ell}\in\mathbb R^r$ be the vectors in head
$\ell$ of the stored matrices $K_j,V_j$.
For $n\geq1$, define
\[
\alpha_{\ell j}(m,a)
=
\frac{
\exp\!\left(\langle q_\ell(a),k_{j,\ell}\rangle/\sqrt r\right)
}{
\displaystyle\sum_{u=1}^n
\exp\!\left(\langle q_\ell(a),k_{u,\ell}\rangle/\sqrt r\right)
}.
\]
Each head normalizes its scores across the stored words.
Its retrieved vector is
\[
y_\ell(m,a)
=
\sum_{j=1}^n\alpha_{\ell j}(m,a)\,v_{j,\ell}.
\]

Choose an output projection $W_O\in\mathbb R^{d\times hr}$.
Concatenating the head outputs gives the word-state transformation
\[
\rho_n(m)(a)
=
W_O\bigl(y_1(m,a)\oplus\cdots\oplus y_h(m,a)\bigr).
\]
Here $\oplus$ denotes concatenation into a vector in
$\mathbb R^{hr}$. At the empty memory, set $\rho_0(*)(a)=0$.
The uncurried attention pairing is
$\widetilde\rho_n(m,a)=\rho_n(m)(a)$.

Thus, a memory determines a function on the word space.
Given a word state, this function constructs its query tensor,
compares each head with the corresponding stored keys, and
combines the stored values into a new word state. This is the
attention computation of Tensor Product Attention
\cite[Section~3.1]{zhang2025tensor}.

The distinction between stored memory and memory representation
is explicit here. Let $E_n$ be the space of lists
$((K_1,V_1),\ldots,(K_n,V_n))$ satisfying
$\operatorname{rank}K_j\leq R_K$ and
$\operatorname{rank}V_j\leq R_V$.
The reconstruction map $\lambda_n:M_n\to E_n$ replaces each
stored factor pair by its tensor. Let $\Psi_n$ denote the
headwise readout defined above. Then
\[
M_n
\xrightarrow{\lambda_n}
E_n
\xrightarrow{\Psi_n}
\operatorname{Hom}(A,A),
\qquad
\rho_n=\Psi_n\circ\lambda_n.
\]
Different factor memories may reconstruct the same tensors
and therefore induce the same transformation.
The resulting family of transformations is
$\mathcal R_n=\rho_n(M_n)$.
With smooth factor encoders, $\mathcal R_n\subseteq C^\infty(A,A)$.

The spaces $E_n$ describe low-rank tensor representations of
the stored information. Their size grows with sequence length;
this construction provides compact representations of individual
keys and values, rather than a fixed-size accumulated memory.

\subsubsection{Layers and causal attention}

Given $s=(a_1,\ldots,a_N)$, first compute its memory $m_s=c_N(s)$.
Applying the represented transformation to every coordinate gives
\[
L_N(s)
=
\bigl(
\rho_N(m_s)(a_1),\ldots,\rho_N(m_s)(a_N)
\bigr).
\]
A feedforward map $f:A\to A$ is then applied independently and
identically at every position. The block operator is
$T_N=f^{\times N}\circ L_N$.

For causal attention, position $i$ uses only its prefix memory:
\[
C_N(s)_i
=
\rho_i\bigl(c_i(a_1,\ldots,a_i)\bigr)(a_i).
\]
The headwise calculation is unchanged, except that both softmax
normalization and value aggregation run over positions
$1,\ldots,i$. Consequently, if $p_i^N:A^N\to A^i$ is the prefix
projection, then
\[
p_i^N\circ C_N=C_i\circ p_i^N.
\]
The causal block
$T_N^{\mathrm{causal}}=f^{\times N}\circ C_N$
inherits this property.

Positional transformations acting on word states may be inserted
at selected layers before the factor encoders are evaluated.
When these transformations are compatible with prefix projections,
their composition with the causal block preserves causality.

\subsubsection{The attention pairing and its memory representation}

Take $\operatorname{Hom}(A,A)=C^\infty(A,A)$. We introduce the attention
pairing as
\[
\rho_n:M_n\longrightarrow\operatorname{Hom}(A,A).
\]
It will factor through the tensor representation:
\[
M_n
\xrightarrow{\lambda_n}
E_n
\xrightarrow{\Psi_n}
\operatorname{Hom}(A,A).
\]

The tensor space carries a headwise contraction
$\kappa:\mathcal T\times\mathcal T\to\mathbb R^h$, defined on
elementary tensors by
\[
\kappa(u\otimes b,\widetilde u\otimes\widetilde b)
=
(u\odot\widetilde u)\langle b,\widetilde b\rangle,
\]
where $\odot$ denotes coordinatewise multiplication.
Bilinear extension gives, in coordinates,
\[
[\kappa(Q,K)]_\ell
=
\sum_{t=1}^r Q_{\ell t}K_{\ell t}.
\]
The contraction retains a separate compatibility score for
each head. Its expression on elementary tensors also permits
these scores to be computed directly from the stored factors.

Let
$e=((K_1,V_1),\ldots,(K_n,V_n))\in E_n$.
For a word state $a\in A$, define the score matrix
$G(e,a)\in\mathbb R^{h\times n}$ by
\[
G(e,a)_{\ell j}
=
\frac{[\kappa(Q(a),K_j)]_\ell}{\sqrt r}.
\]
For $n\geq1$, apply softmax separately across the stored words
within each head:
\[
P(e,a)_{\ell j}
=
\frac{\exp(G(e,a)_{\ell j})}
{\displaystyle\sum_{u=1}^n\exp(G(e,a)_{\ell u})}.
\]
The retrieved tensor $Y(e,a)\in\mathcal T$ is
\[
Y(e,a)_{\ell t}
=
\sum_{j=1}^n P(e,a)_{\ell j}(V_j)_{\ell t}.
\]
The output tensor need not satisfy the rank bound imposed on
the individual value tensors.

Choose an output projection
$W_O\in\mathbb R^{d\times hr}$.
Let $\operatorname{vec}_{\mathrm{head}}(Y)\in\mathbb R^{hr}$
denote the vector obtained by concatenating the rows of $Y$,
one head at a time. Define
\[
\Psi_n(e)(a)
=
W_O\operatorname{vec}_{\mathrm{head}}(Y(e,a)).
\]
The attention pairing is therefore
\[
\rho_n=\Psi_n\circ\lambda_n,
\qquad
\widetilde\rho_n(m,a)=\rho_n(m)(a).
\]
At the empty memory, set $\rho_0(*)(a)=0$.

The associated memory representation is
\[
\mathcal R_n
=
\rho_n(M_n)
=
\Psi_n(\lambda_n(M_n))
\subseteq C(A,A).
\]
With the smooth factor maps chosen above, its elements are
smooth self-maps of $A$. Distinct factor memories that
reconstruct the same keys and values induce the same
transformation. Thus, the stored factors, their tensor
representations, and the resulting word-state transformations
are distinct levels of the construction.

The bilinear map $\kappa$ supplies the headwise compatibility
scores. The full attention pairing additionally includes
query encoding, softmax normalization, value aggregation,
and the output projection, and is generally nonlinear.

\subsubsection{Attention layers and causal computation}

For $s=(a_1,\ldots,a_N)$, its total memory $m_s=c_N(s)$
determines the attention operator
\[
L_N:A^N\longrightarrow A^N,
\qquad
L_N(s)
=
\bigl(
\rho_N(m_s)(a_1),\ldots,\rho_N(m_s)(a_N)
\bigr).
\]
For each word, the query tensor is compared with all stored
key tensors. Each head then aggregates the corresponding
rows of the value tensors, and the concatenated result is
projected back into $A$. This realizes the tensor-factorized
attention computation of
\cite[Section~3.1]{zhang2025tensor}.

A feedforward map $f:A\to A$ is applied independently and
identically at each position. The resulting block is
\[
T_N=f^{\times N}\circ L_N.
\]

For causal attention, position $i$ uses only its prefix memory
$m_i=c_i(a_1,\ldots,a_i)$. Define
\[
C_N(s)_i=\rho_i(m_i)(a_i).
\]
In this computation, the score matrix contains only the first
$i$ keys, softmax normalizes over those $i$ positions within
each head, and aggregation uses only their associated values.
Consequently, for the prefix projection $p_i^N:A^N\to A^i$,
\[
p_i^N\circ C_N=C_i\circ p_i^N.
\]
The causal attention and feedforward block
$T_N^{\mathrm{causal}}=f^{\times N}\circ C_N$
inherits the same prefix compatibility.

Positional transformations acting on word states may be
included at selected layers by applying
$\Phi_N=\phi_1\times\cdots\times\phi_N$ before constructing
the query, key, and value factors. When these transformations
are compatible with prefix projections, their composition
with the causal block preserves causality.

\subsection{Intention as regression memory}\label{intention}

Intention constructs a transformation from memory by fitting a regression
function to the observed key--value associations
\cite{garnelo2023exploring}. In our framework, its defining map
$\rho_n:M_n\to\operatorname{Hom}(A,A)$ therefore represents memory through
a fitted operator. We describe the regularized least-squares form of
this construction.

Let $A=\mathbb R^d$, and fix continuous feature maps
$q,k:A\to\mathbb R^r$ and $v:A\to\mathbb R^p$, together with a linear
output map $W_O:\mathbb R^p\to A$. For a sequence
$s=(a_1,\ldots,a_n)$, write
\[
K_n=[k(a_1)\mid\cdots\mid k(a_n)],
\qquad
V_n=[v(a_1)\mid\cdots\mid v(a_n)].
\]
The key--value memory $(K_n,V_n)$ supplies the data for a regression
problem: we seek a function $F:\mathbb R^r\to\mathbb R^p$ such that
$F(K_n)\approx V_n$, where $F(K_n)$ means applying $F$ to each column
of $K_n$. Once fitted, this function acts on the query features of
each word.

For linear regression with regularization parameter $\gamma>0$,
write $F_n(\xi)=B_n\xi$. The fitted matrix is
\[
B_n
=
\operatorname*{arg\,min}_{B\in\mathbb R^{p\times r}}
\left\{
\|BK_n-V_n\|_F^2+\gamma\|B\|_F^2
\right\}
=
V_nK_n^\top(K_nK_n^\top+\gamma I_r)^{-1}.
\]
Thus the fitting procedure depends on the key--value memory only
through the two accumulated statistics
$G_n=K_nK_n^\top$ and $S_n=K_nV_n^\top$.

We take $M_n=(\mathbb R^r\times\mathbb R^p)^n$, identifying
its elements with the key--value arrays $(K,V)$. Starting from the
empty arrays, the memory update is
\[
\mu_n((K,V),a)
=
([K\mid k(a)],[V\mid v(a)]).
\]
The resulting compression map is $c_n(s)=(K_n,V_n)$. The accumulated
statistics belong to the matrix space
\[
E_n=
\left\{
[G\mid S]:
G\in\operatorname{Sym}_r^{\geq0},\quad
S\in\mathbb R^{r\times p},\quad
\operatorname{rank}[G\mid S]=\operatorname{rank}G\leq n
\right\},
\]
where $\operatorname{Sym}_r^{\geq0}$ denotes the positive semidefinite
symmetric matrices. Their representation and update are
\[
\begin{aligned}
\lambda_n(K,V)&=[KK^\top\mid KV^\top],\\
\bar\mu_n([G\mid S],a)
&=[G+k(a)k(a)^\top\mid S+k(a)v(a)^\top].
\end{aligned}
\]
This matrix representation has a fixed number of coordinates,
independently of the sequence length. Distinct key--value memories
can have the same matrix representation.

To distinguish memory from its represented transformation, set
$E=\operatorname{Hom}_{\mathbb R}(\mathbb R^r,\mathbb R^p)$.
A matrix representation $H=[G\mid S]$ determines the regression operator
\[
\eta_n(H)=F_H\in E,
\qquad
F_H(\xi)=S^\top(G+\gamma I_r)^{-1}\xi.
\]
For a memory obtained from a sequence, $F_{\lambda_n(c_n(s))}$ is precisely
the fitted function $F_n$ above. Define the decoder by
$\Psi(F)(a)=W_OF(q(a))$. The memory representation then factors as
\[
\begin{gathered}
M_n\xrightarrow{\lambda_n}E_n\xrightarrow{\eta_n}E
\xrightarrow{\Psi}\operatorname{Hom}(A,A),\\
\rho_n=\Psi\circ\eta_n\circ\lambda_n.
\end{gathered}
\]
Equivalently, the attention pairing is
\[
\widetilde\rho_n(m,a)
=
W_OS^\top(G+\gamma I_r)^{-1}q(a),
\qquad [G\mid S]=\lambda_n(m).
\]
Although $E$ consists of linear operators on query features, the
induced word transformation need not be linear on $A$, since $q$
may be nonlinear.

This construction gives a precise meaning to \emph{regression memory}.
The accumulated state $(G,S)$ retains the information needed to fit
and update a regression rule, while $F_H$ represents the rule currently
determined by that state. These roles are distinct: different memories
can determine the same fitted operator but yield different operators
after receiving another word. Hence the fitted operator alone is
generally insufficient for subsequent memory updates.

The attention layer applies the fitted function to every query:
\[
L_N(s)_i
=
W_OF_{\lambda_N(c_N(s))}(q(a_i)),
\qquad
T_N=f^{\times N}\circ L_N,
\]
where $f:A\to A$ is the feedforward map. In the causal construction,
each position uses the regression function fitted from its own prefix:
\[
C_N(s)_i
=
W_OF_{\lambda_i(c_i(s_{\leq i}))}(q(a_i)).
\]
The corresponding attention-plus-feedforward block is
$f^{\times N}\circ C_N$.

\subsection{Recurrent neural networks and direct memory readout}
\label{rnn}

Recurrent neural networks use the accumulated memory itself as the
contextual representation from which predictions are made. In our
framework, this corresponds to equipping each memory space $M_n$
with a readout map $r_n:M_n\to Y$, where $Y$ is the output space.

The memory is constructed through the updates
$\mu_n:M_n\times A\to M_{n+1}$ introduced above. Starting from an
initial state $h_0\in M_0$, a sequence $(a_1,\ldots,a_n)$ determines
\[
h_n=\mu_{n-1}(h_{n-1},a_n)
=c_n(a_1,\ldots,a_n).
\]
The prediction is then obtained directly from this state:
\[
A^n\xrightarrow{c_n}M_n\xrightarrow{r_n}Y,
\qquad
\widehat y_n=r_n(h_n).
\]
Thus the memory state summarizes the processed sequence, while the
readout converts that summary into the desired prediction. For
next-word prediction, $Y$ may be taken to be the probability simplex
over the vocabulary.

A standard RNN uses a fixed memory space $M_n=M$ and shares the
same update $\mu:M\times A\to M$ across positions. For example,
with $A=\mathbb R^d$ and $M=\mathbb R^m$, a vanilla RNN has
\[
h_n
=
\sigma(W_hh_{n-1}+W_a a_n+b),
\qquad
\widehat y_n=r(h_n),
\]
where $W_h\in\mathbb R^{m\times m}$,
$W_a\in\mathbb R^{m\times d}$, and $\sigma$ is an activation
function applied coordinatewise. The readout $r$ may include an
output projection and softmax. The word-state space $A$ and the
memory space $M$ need not coincide.

The same construction supports a prediction at every position.
For $s=(a_1,\ldots,a_N)$, define
\[
R_N(s)_i=r_i\bigl(c_i(s_{\leq i})\bigr),
\qquad 1\leq i\leq N.
\]
Each output depends only on the corresponding prefix, so these
sequence maps are causal: extending the input sequence does not
change the outputs already determined by its earlier prefixes.

The relation to attention lies in how the accumulated memory is
read. An attention pairing assigns to each memory a transformation
of word states,
\[
\rho_n:M_n\longrightarrow\operatorname{Hom}(A,A),
\qquad
\widetilde\rho_n(m,a)=\rho_n(m)(a).
\]
Its output depends on both the memory and the word state at which
the represented transformation is evaluated. An RNN readout
$r_n:M_n\to Y$, by comparison, uses the memory directly as the
representation for prediction.

Recurrence and attention therefore describe different aspects of
the computation. The maps $\mu_n$ specify how memory develops as
words are processed; the maps $r_n$ or $\rho_n$ specify how that
memory is used. In particular, the memories of linear attention
and Intention admit recurrent updates, while their attention
pairings read those memories through a query. Our framework
accommodates both direct memory readout and word-conditioned
memory readout within the same construction of memory.

\section{Geometric spaces and architectural design}

The preceding examples show how architectural choices can be expressed
through word states, memory updates, and memory readouts. The memory
spaces $M_n$ specify which states can be accumulated, while the maps
$\mu_n:M_n\times A\to M_{n+1}$ determine how these states change as
words are processed. For attention, the map
$\rho_n:M_n\to\operatorname{Hom}(A,A)$ specifies the transformation
of word states represented by each memory. For recurrent prediction,
a readout $r_n:M_n\to Y$ acts directly on the memory.

Tables~\ref{tab:knowledge-intelligence-architectures}
and~\ref{tab:memory-readouts} summarize these choices for the
architectures discussed above. In the update formulas, $k$ and $v$
abbreviate $k(a)$ and $v(a)$; the feature maps and output projections
are those of the corresponding architecture. For tensor product
attention, $e_K(a)$ and $e_V(a)$ denote the stored factor pairs from
Section~\ref{tensor attention}.

\begin{table}[htbp]
\centering
\footnotesize
\setlength{\tabcolsep}{4pt}
\renewcommand{\arraystretch}{1.3}
\begin{tabularx}{\textwidth}{
    @{}
    >{\raggedright\arraybackslash}p{0.17\textwidth}
    *{2}{>{\raggedright\arraybackslash}X}
    @{}
}
\toprule
\textbf{Architecture}
&
\textbf{Memory and representation}
&
\textbf{Update on receiving $a$}
\\
\midrule

Softmax attention
&
Key and value arrays $(K,V)$.
&
$(K,V)\mapsto([K\mid k],[V\mid v])$.
\\
\midrule

Linear attention
&
Ordered pairs $x=(k_j,v_j)_{j=1}^n\in M_n$.
\newline
$\lambda_n(x)=[z\mid S]\in E_n$, where
\newline
$z=\sum_j k_j$ and $S=\sum_j k_jv_j^\top$,
with $\operatorname{rank}[z\mid S]\le n$.
&
Append $(k,v)$ to $x$.
\newline
Encoded update $\bar\mu_n$:
\newline
$(z,S)\mapsto(z+k,S+kv^\top)$.
\\
\midrule

Tensor product attention
&
A sequence of key and value factor pairs
$m=(e_{K,j},e_{V,j})_{j=1}^n$.
&
Append $(e_K(a),e_V(a))$ to $m$.
\\
\midrule

Intention
&
Ordered pairs $x=(k_j,v_j)_{j=1}^n\in M_n$.
\newline
$\lambda_n(x)=[G\mid S]\in E_n$, where
\newline
$G=\sum_j k_jk_j^\top$ and $S=\sum_j k_jv_j^\top$,
with $\operatorname{rank}[G\mid S]\le n$.
&
Append $(k,v)$ to $x$.
\newline
Encoded update $\bar\mu_n$:
\newline
$(G,S)\mapsto(G+kk^\top,S+kv^\top)$.
\\
\midrule

RNN
&
A hidden state $h\in M$.
&
$h\mapsto u_\theta(h,a)$, where $u_\theta$ is the recurrent update.
\\

\bottomrule
\end{tabularx}
\caption{Memory states and sequential updates. Linear attention,
Intention, and a fixed-width RNN use a fixed-dimensional running
representation. Softmax attention and tensor product attention retain
a sequence of records in the realizations shown here.}
\label{tab:knowledge-intelligence-architectures}
\end{table}

The representation of memory specifies how its stored information
becomes a transformation. An attention pairing may factor as
\[
M_n\xrightarrow{\lambda_n}E_n
\xrightarrow{\Psi_n}\operatorname{Hom}(A,A),
\qquad
\rho_n=\Psi_n\circ\lambda_n.
\]
Here $E_n$ is a chosen carrier for the represented transformations;
it need not coincide with their image in $\operatorname{Hom}(A,A)$.
A common finite-dimensional space $E$ and decoder $\Psi$ for all
$n$ provide a representation whose dimension does not grow with
sequence length. In the preceding linear attention and Intention
constructions, $\Psi_n=\Psi\circ\eta_n$.

We call the memory representation faithful on reachable memories if
$\rho_n$ is injective on $c_n(A^n)$. In linear attention and Intention,
distinct ordered memories may have the same $\lambda_n$-image and
hence the same represented transformation. Their recurrent implementations
retain this matrix representation rather than the full ordered memory.

\begin{table}[htbp]
\centering
\footnotesize
\setlength{\tabcolsep}{4pt}
\renewcommand{\arraystretch}{1.3}
\begin{tabularx}{\textwidth}{
    @{}
    >{\raggedright\arraybackslash}p{0.17\textwidth}
    *{2}{>{\raggedright\arraybackslash}X}
    @{}
}
\toprule
\textbf{Architecture}
&
\textbf{Representation used for readout}
&
\textbf{Resulting output}
\\
\midrule

Softmax attention
&
Retain $(K,V)$ in a carrier $E_n$ whose size grows with $n$.
&
$W_OV\alpha$, where
\newline
$\alpha=\operatorname{softmax}\!
\left(K^\top q(a)/\sqrt r\right)$.
\\
\midrule

Linear attention
&
The linear operator
\newline
$\ell_{z,S}(\xi)=(S^\top\xi,z^\top\xi)$
on query features.
&
$\displaystyle
W_O\frac{S^\top q(a)}{\varepsilon+z^\top q(a)}$,
\newline
with $\varepsilon>0$.
\\
\midrule

Tensor product attention
&
Reconstruct $(K_j,V_j)_{j=1}^n$ from their stored factors.
&
Headwise softmax attention, followed by concatenation and $W_O$.
\\
\midrule

Intention
&
The regression operator
\newline
$F_{G,S}(\xi)$
\newline
$\quad=S^\top(G+\gamma I)^{-1}\xi$,
\newline
with $\gamma>0$.
&
$W_OF_{G,S}(q(a))$.
\\
\midrule

RNN
&
The hidden state $h$ itself.
&
$r(h)\in Y$.
\\

\bottomrule
\end{tabularx}
\caption{Memory representations and readouts. The attention rows
give $\rho_n(m)(a)$; the RNN row gives a direct prediction from
memory. Linear attention applies normalization after evaluating
its represented linear operator. Intention represents a fitted
linear map on query features.}
\label{tab:memory-readouts}
\end{table}

For the attention architectures, the displayed word transformation
is applied at each position, followed by the pointwise feedforward
map $f^{\times N}$. Full attention uses the total memory $c_N(s)$,
whereas causal attention uses $c_i(s_{\leq i})$ at position $i$.
A recurrent predictor similarly reads each prefix memory when
producing a sequence of outputs.

These examples suggest a design strategy in which geometry guides
both the construction and the use of memory. One first chooses
word-state and memory spaces that reflect relevant structure in
the data, then defines compatible updates and a family of represented
transformations. For example, if the task requires an order-independent
memory representation, the encoding should satisfy
$\lambda_n\circ c_n\circ\pi=\lambda_n\circ c_n$ for every
coordinate permutation $\pi$ of $A^n$. Adding position-independent
word contributions in a vector space is one way to obtain this
property.

Choosing a memory space alone does not determine an attention
layer. One must also specify how words update that space and how
the resulting memory produces an endomorphism of $A$. A scalar
pairing can contribute to this construction, but a complete
attention pairing must return a word state. The choice of $E_n$
and $\Psi_n$ makes explicit which transformations the memory can
represent and how those transformations are evaluated.

A compact representation of the current transformation need not
contain enough information to update it after another word arrives.
If the representation spaces $E_n$ are to serve as the running
memory, their updates $\bar\mu_n:E_n\times A\to E_{n+1}$ must satisfy
\[
\lambda_{n+1}\circ\mu_n
=
\bar\mu_n\circ(\lambda_n\times\operatorname{id}_A).
\]
Intention illustrates this distinction: its fitted regression
operator determines the current readout, while further updates
generally require retaining covariance information as well.

The design process can therefore be organized as follows:
\begin{center}
\resizebox{\linewidth}{!}{%
\begin{tikzpicture}[
    node distance=7mm,
    stage/.style={
        draw=black!60,
        fill=black!3,
        rounded corners=3pt,
        line width=0.6pt,
        minimum height=16mm,
        text width=28mm,
        inner sep=5pt,
        align=center
    },
    flow/.style={
        -{Stealth[length=2.5mm]},
        line width=0.8pt,
        draw=black!70
    }
]
    \node[stage] (data)
        {Data structure\\and task};

    \node[stage, right=of data, text width=38mm] (memory)
        {Word states, memory\\spaces, and updates};

    \node[stage, right=of memory, text width=38mm] (readout)
        {Memory representations\\and readouts};

    \node[stage, right=of readout] (model)
        {Layer composition\\and evaluation};

    \draw[flow] (data) -- (memory);
    \draw[flow] (memory) -- (readout);
    \draw[flow] (readout) -- (model);
\end{tikzpicture}%
}
\end{center}

The geometric constructions below provide starting points for this
process. Their role is to suggest memory spaces and structured
families of transformations from which compatible updates and
readouts can be developed. The computational cost and predictive
usefulness of the resulting layers must then be evaluated for
the intended task.

\subsection{Manifold embedding}
\label{geometric constructions}

A direct way to introduce geometry into attention is to map
queries and keys into a manifold and compare their embedded
representations by an inner product. This changes the
query--key similarity while retaining the value aggregation
and feedforward composition described above.

Let $\mathcal N$ be a smooth manifold equipped with an embedding
$\iota:\mathcal N\hookrightarrow F$, where $F$ is a
finite-dimensional inner-product space. Choose a map
$\chi:\mathbb R^r\to\mathcal N$ and write
$\phi=\iota\circ\chi$. For query and key coordinates
$u,w\in\mathbb R^r$, the resulting similarity is
$\kappa_\phi(u,w)=\langle\phi(u),\phi(w)\rangle_F$.
Thus the inner product is evaluated in the ambient space $F$;
the manifold and its embedding determine which similarities
are available.

Let $q,k:A\to\mathbb R^r$ be the query and key maps,
$v:A\to\mathbb R^p$ the value map, and
$W_O:\mathbb R^p\to A$ the output projection.
For an ordered memory
$x=((k_1,v_1),\ldots,(k_n,v_n))\in M_n$ and a scale
$\beta>0$, define the attention pairing by
\[
\rho_n(x)(a)
=
W_O
\frac{
\displaystyle\sum_{j=1}^n
\exp\!\bigl(\beta\kappa_\phi(q(a),k_j)\bigr)v_j
}{
\displaystyle\sum_{j=1}^n
\exp\!\bigl(\beta\kappa_\phi(q(a),k_j)\bigr)
},
\qquad n\geq1.
\]
At the empty memory, take $\rho_0$ to be the zero transformation.
This is softmax attention with the similarity determined by
the chosen manifold embedding.

The corresponding representation space can be taken as
$E_n=(\iota(\mathcal N)\times\mathbb R^p)^n$, with
$\lambda_n(x)=((\phi(k_j),v_j))_{j=1}^n$.
The decoder evaluates the embedded query against these
embedded keys, giving the factorization
$\rho_n=\Psi_n\circ\lambda_n$.
Here the geometric choice concerns the representation of
each key; the ordered memory still contains its individual
key--value records.

\noindent\textbf{Spherical embeddings.}
Take $\mathcal N=S^r\subset\mathbb R^{r+1}$ and let
$\iota$ be the inclusion. An inverse stereographic map
with scale $c>0$ is
\[
\phi_{\mathrm{sph}}(u)
=
\left(
\frac{2cu}{c^2+\|u\|^2},
\frac{c^2-\|u\|^2}{c^2+\|u\|^2}
\right).
\]
Its image lies on the unit sphere, so queries and keys
are compared by the ordinary Euclidean inner product
of their spherical images. This map is injective onto
the sphere with one point removed. In particular,
the magnitude of the original vector remains encoded
in its position on the sphere. The choice $c=2$ gives
the feature map used in our spherical construction.

\noindent\textbf{Projective embeddings.}
Take $\mathcal N=\mathbb{RP}^r$ and map a coordinate
vector $u$ to the projective point $[u:1]$.
A canonical matrix representation is obtained by setting
$\widehat u=(u,1)/\sqrt{1+\|u\|^2}$ and
$P_u=\widehat u\widehat u^\top$.
The projector depends only on the projective point,
rather than on its homogeneous representative.

Using the centered embedding
$\phi_{\mathrm{proj}}(u)=P_u-I_{r+1}/(r+1)$ into the
space of traceless symmetric matrices, equipped with
the Frobenius inner product, gives
\[
\kappa_{\mathrm{proj}}(u,w)
=
\operatorname{tr}(P_uP_w)-\frac1{r+1}
=
\frac{(1+u^\top w)^2}
     {(1+\|u\|^2)(1+\|w\|^2)}
-\frac1{r+1}.
\]
This is the homogeneous projective similarity used in
our corresponding attention construction. The map
$u\mapsto[u:1]$ is injective: projective equivalence
acts on the entire homogeneous representative and
does not identify distinct original coordinate vectors.

\noindent\textbf{Grassmannian embeddings.}
Take $\mathcal N=\operatorname{Gr}(s,D)$, the manifold
of $s$-dimensional subspaces of $\mathbb R^D$.
When $r=s(D-s)$, a query or key vector can be reshaped
into a matrix $X\in\mathbb R^{(D-s)\times s}$.
The columns of $A_X=[X^\top\mid I_s]^\top$ span its
graph, giving a subspace with orthogonal projector
\[
P_X=A_X(I_s+X^\top X)^{-1}A_X^\top,
\qquad
\phi_{\mathrm{Gr}}(X)=P_X-\frac{s}{D}I_D.
\]
The graph matrix has full column rank for every $X$,
and the resulting projector is independent of the
basis chosen for that subspace.

With the Frobenius inner product, the similarity is
$\kappa_{\mathrm{Gr}}(X,Y)
=\operatorname{tr}(P_XP_Y)-s^2/D$.
If $U_X$ and $U_Y$ are orthonormal bases, then
$\operatorname{tr}(P_XP_Y)=\|U_X^\top U_Y\|_F^2$,
the sum of the squared cosines of the principal angles
between the subspaces. Thus the score compares subspaces
through their projector embeddings. Our Grassmannian
construction uses $\operatorname{Gr}(4,12)$, mapping
32 query or key coordinates into a 32-dimensional
manifold represented in the 77-dimensional space of
traceless symmetric matrices.

In the projective and Grassmannian constructions,
centering subtracts the same constant from every
query--key score. At a fixed scale $\beta$, this
does not change the softmax weights. It provides a
convenient ambient representation for describing
the geometry.

These examples distinguish the intrinsic dimension
of a manifold from the dimension of its ambient
feature space. They also show that introducing a
manifold need not discard the original query or
key coordinates: the stereographic, homogeneous,
and graph maps above are injective on their
coordinate domains. Their architectural effect
comes from the similarities induced by the chosen
embeddings. Faithfulness of the complete map
$\rho_n:M_n\to\operatorname{Hom}(A,A)$ remains a
separate question.

The attention layer is obtained by applying the
represented transformation to each word, followed
by the pointwise feedforward map as before.
The same construction applies headwise in a
multihead layer. The spherical, projective, and
Grassmannian choices therefore provide different
geometric query--key comparisons within the same
general attention architecture.
\subsection{Query actions and value readout}

Another geometric construction treats queries as transformations
acting on keys, with values providing the subsequent readout.
The interaction therefore takes the form
\[
K_j \xmapsto{\ Q_i\ } Q_i(K_j)
    \xmapsto{\ V_j\ } V_j\bigl(Q_i(K_j)\bigr).
\]
The intermediate response may have several coordinates,
allowing a query to interact with the same key through
multiple actions.

\noindent\textbf{Queries as transformations.}
Let $\mathcal K$ be a key space,
$\mathcal U=\mathbb R^t$ a response space, and
$\mathcal W=\mathbb R^p$ an output space.
Choose a family
$\mathcal Q\subseteq\operatorname{Hom}(\mathcal K,\mathcal U)$
of query transformations. For $A=\mathbb R^d$, the layer
has encoders $q:A\to\mathcal Q$, $k:A\to\mathcal K$, and
$v:A\to\operatorname{Hom}_{\mathbb R}(\mathcal U,\mathcal W)$,
together with an output projection $W_O:\mathcal W\to A$.

For a sequence $(a_1,\ldots,a_N)$, write
$Q_i=q(a_i)$, $K_j=k(a_j)$, and $V_j=v(a_j)$.
The query $Q_i$ acts on $K_j$ to produce
$Q_i(K_j)\in\mathcal U$.
The value $V_j$ is a linear map that converts a response
in $\mathcal U$ into an output in $\mathcal W$.

\noindent\textbf{Normalization and attention pairing.}
The ordered memory space is
$M_n=(\mathcal K\times
\operatorname{Hom}_{\mathbb R}(\mathcal U,\mathcal W))^n$,
with memory formation appending the computed pair
$(k(a),v(a))$.
For $m=((K_1,V_1),\ldots,(K_n,V_n))$,
normalize each response coordinate over the keys.
Using softmax with scale $\beta>0$, define
$w_j(Q;m)\in\mathbb R^t$ by
\[
[w_j(Q;m)]_b
=
\frac{\exp\!\bigl(\beta[Q(K_j)]_b\bigr)}
     {\displaystyle\sum_{\ell=1}^{n}
      \exp\!\bigl(\beta[Q(K_\ell)]_b\bigr)},
\qquad b=1,\ldots,t.
\]
Thus $\sum_{j=1}^{n}[w_j(Q;m)]_b=1$ for each
response coordinate $b$. Each action has its own
distribution over the same collection of keys.

The memory induces the transformation
$\rho_n(m)\in\operatorname{Hom}(A,A)$ given by
\[
\rho_n(m)(a)
=
W_O\sum_{j=1}^{n}
V_j\bigl(w_j(q(a);m)\bigr),
\qquad n\geq1.
\]
At the empty memory, take $\rho_0=0$.
This realizes the attention pairing by three operations:
the query acts on the keys, the responses are normalized
across memory records, and the values map those responses
into outputs that are added together.

\noindent\textbf{Linear actions.}
When $\mathcal K=\mathbb R^r$ and the query
transformations are linear, a query is represented by
a matrix $Q_i\in\mathbb R^{t\times r}$, while a value
is represented by $V_j\in\mathbb R^{p\times t}$.
The response is $Q_iK_j$, and the value readout is
ordinary matrix multiplication.

For $t=2$, write
\[
Q_i=
\begin{pmatrix}
q_{i,1}^{\top}\\
q_{i,2}^{\top}
\end{pmatrix},
\qquad
V_j=
\begin{pmatrix}
v_{j,1}&v_{j,2}
\end{pmatrix}.
\]
Both query rows act on the same key $K_j$.
Writing $w_{ij}=w_j(Q_i;m)$, the contribution of
record $j$ is
$V_jw_{ij}=w_{ij,1}v_{j,1}+w_{ij,2}v_{j,2}$.
The two actions therefore share the key representation,
while retaining separate query transformations,
normalizations, and value outputs. The same construction
allows any response dimension $t$.

\noindent\textbf{Relation to scalar attention.}
Ordinary dot-product attention is recovered when $t=1$:
the query is a linear functional on the key space,
and a value vector represents the linear map
$c\mapsto cv_j$ from $\mathbb R$ to $\mathcal W$.
The manifold construction above also fits this scalar
case by taking the query transformation associated
with $u$ to be
$Q_u(w)=\langle\phi(u),\phi(w)\rangle_F$.

More generally, the query transformations may be
nonlinear maps on a geometric key space. Choosing
the key space, the family of transformations, and
the response dimension determines how queries act
on memory, while the value maps determine how
those responses become word-state updates.
Applying the resulting transformation at each
position and composing with the pointwise
feedforward map gives the layer as before.

\subsection{Further geometric and functional constructions}

Further possibilities arise by changing the token space,
the memory representation, or the operation through which
memory updates token states. Throughout these constructions,
$M_n$ retains the ordered encoded records, while aggregate
matrices and functions belong to representation spaces $E_n$.
Scalar pairings can be combined with value readouts to obtain
the transformation $\rho_n(m):A\to A$.

\noindent\textbf{Lie-algebra states and exponential updates.}
Tokens may be encoded as matrices in a real matrix Lie algebra
$\mathfrak g$, whose bracket is $[X,Y]=XY-YX$.
Closure under this operation allows interactions and updates
to remain in the same algebra. For example, encoded keys and
values can determine
$H_i=\sum_j\alpha_{ij}[K_j,V_j]\in\mathfrak g$,
where $\alpha_{ij}$ are scalar routing weights.
A token state $X_i\in\mathfrak g$ can then be updated through
the bracket or its exponential counterpart:
\[
X_i^{+}=X_i+\eta[H_i,X_i],
\qquad\text{or}\qquad
X_i^{+}=e^{\eta H_i}X_i e^{-\eta H_i}.
\]
For fixed $H_i$, the first is an Euler step for the bracket
flow and the second is its exact evolution. The exponential
lies in the associated matrix group, and conjugation keeps
the token on its adjoint orbit. In particular,
$\mathfrak{so}(d)$ gives skew-symmetric states acted on by
orthogonal transformations.

\noindent\textbf{Spline representations of memory.}
An ordered memory can supply the control coefficients of a
function that queries subsequently evaluate. For
$m=((k_1,v_1),\ldots,(k_n,v_n))$, with
$k_j,v_j\in\mathbb R^p$, arrange the coefficients cyclically
as $(k_1,v_1,\ldots,k_n,v_n)$.
Let $f_{m,c}$ be the periodic cubic B-spline of period $2n$
determined by coordinate $c$ of these coefficients.
A query map $q:A\to\mathbb R^p$ selects a separate address
for each coordinate, giving
\[
\rho_n(m)(a)
=
W_O\left(
f_{m,c}\bigl(2n\,\sigma(q_c(a))\bigr)
\right)_{c=1}^{p},
\qquad n\geq1,
\]
where $\sigma$ is the sigmoid function.
Each scalar evaluation uses at most four neighboring control
coefficients. Here $\lambda_n$ represents memory by a family
of spline functions, and the query determines where those
functions are evaluated. The ordering of the records
determines their arrangement along the cyclic domain.

\noindent\textbf{Symmetric and skew-symmetric representations.}
Structured matrix increments provide rank-filtered
representations of memory. Choosing
$B_j=c_j u_j u_j^\top$ gives a symmetric aggregate
$S_n=\sum_j B_j$ with $\operatorname{rank}S_n\leq n$,
while choosing $B_j=u_jv_j^\top-v_ju_j^\top$ gives
$H_n=\sum_j B_j\in\mathfrak{so}(d)$ with
$\operatorname{rank}H_n\leq2n$.
These aggregates define $\lambda_n$, and appending a record
adds its corresponding increment.
The trace pairings $\operatorname{tr}(S_nB)$ and
$-\tfrac12\operatorname{tr}(H_nB)$ evaluate symmetric and
skew-symmetric queries, respectively, supplying scalar
responses for a subsequent readout.

\noindent\textbf{Special linear representations.}
The traceless algebra $\mathfrak{sl}(d,\mathbb R)$ admits
rank-one increments $B_j=u_jv_j^\top$ satisfying
$v_j^\top u_j=0$. Their sums define representations in
$E_n=\{H\in\mathfrak{sl}(d,\mathbb R):
\operatorname{rank}H\leq n\}$, with evaluation through
$\operatorname{tr}(HB)$ against a traceless query matrix.
Each increment satisfies $B_j^2=0$, so
$\exp(B_j)=I+B_j$.
More generally, exponentiating a traceless aggregate produces
a matrix of determinant one. This supplies both additive
representations in the algebra and multiplicative updates
in the associated special linear group.

\noindent\textbf{Symplectic representations.}
Let $J=\left(\begin{smallmatrix}0&I_d\\-I_d&0\end{smallmatrix}\right)$.
The symplectic Lie algebra consists of matrices satisfying
$H^\top J+JH=0$ and admits rank-one increments
$B_j=c_j u_j u_j^\top J$.
Their aggregate $H_n=\sum_jB_j$ remains in
$\mathfrak{sp}(2d,\mathbb R)$ and has rank at most $n$,
giving a corresponding rank-filtered representation space.
A scalar readout is supplied by $-\operatorname{tr}(H_nB)$.
Alternatively, $G=\exp(H_n)$ satisfies $G^\top JG=J$,
so its action preserves the chosen symplectic form.
This makes preservation of a bilinear geometric structure
an explicit constraint on the update.

\noindent\textbf{Polynomial representations of ordered memory.}
For records $x_j\in\mathbb R^d$, define
$P_m(z)=\prod_{j=1}^{n}(1+x_j^\top z)-1$.
The map $\lambda_n:m\mapsto P_m$ takes ordered memory into
polynomials of degree at most $n$ and is invariant under
permutations of its records.
Appending $x$ induces the update
$P_{m\cdot x}(z)=(P_m(z)+1)(1+x^\top z)-1$.
Evaluation at a query produces a scalar response containing
products of contributions from multiple records.
Thus permutation invariance is a property of the
representation and readout, while $M_n$ remains an
ordinary product.

\noindent\textbf{Projective polynomial representations.}
For ordered records $([u_1],\ldots,[u_n])\in(\mathbb{CP}^1)^n$,
symmetric multiplication gives the representation
$[P]=[u_1\odot\cdots\odot u_n]$ in
$E_n=\mathbb P(\operatorname{Sym}^n\mathbb C^2)$.
Appending $[u]$ updates the representation by
$[P]\mapsto[P\odot u]$.
Using the induced Hermitian inner product, a projective
query $[q]\in\mathbb{CP}^1$ can be evaluated by
\[
\kappa_n([P],[q])
=
\frac{|\langle P,q^{\otimes n}\rangle|^2}
     {\|P\|^2\|q\|^{2n}}.
\]
This expression is independent of the homogeneous
representatives. The degree grows with memory length,
representing collective memory through a homogeneous
polynomial whose evaluation can guide the word-state readout.

\section{Geometric principles for improving attention layers}
\label{geometric improvement}

\subsection{Memory faithfulness, storage, and evaluation cost}
\label{faithfulness and cost}

The computational cost of attention raises a geometric question:
which distinctions between memories can a layer preserve, how much
space does it take to store them, and how many operations does it
take to act on a word? Recall the factorization
\[
M_n \xrightarrow{\lambda_n} E_n
    \xrightarrow{\Psi_n} \operatorname{Hom}(A,A),
\qquad
\rho_n=\Psi_n\circ\lambda_n.
\]
Here $M_n$ records the ordered key--value pairs computed by the
layer, while $\lambda_n(m)$ is the representation used for
evaluation. The faithfulness of $\rho_n$ determines which memories
can be distinguished through their actions on words. If two
memories induce the same transformation, no query to this layer
can distinguish them. Such an identification limits performance
when the task requires different responses to the same query.

This observation gives a precise sense in which a larger
representation may be necessary. Suppose a reachable family of
memories is parametrized by an open subset of $\mathbb R^D$, and
all its members must induce distinct responses. A continuous
representation of this family in $\mathbb R^s$ must then have
$s\geq D$, by invariance of domain. Storage and evaluation,
however, are different costs: the former depends on how
$\lambda_n(m)$ is encoded, while the latter depends on the
operations required to compute $\Psi_n(\lambda_n(m))(a)$.
Softmax and linear attention illustrate this distinction.

\noindent\textbf{Softmax attention: a response family that grows
with memory length.}
Consider one head with keys $k_j\in\mathbb R^r$, values
$v_j\in\mathbb R^p$, and query domain $Q\subseteq\mathbb R^r$.
Before the output projection, its memory represents the function
\[
F_m(\xi)=
\frac{\sum_{j=1}^{n}\exp(\beta\,\xi^\top k_j)v_j}
     {\sum_{j=1}^{n}\exp(\beta\,\xi^\top k_j)},
\qquad \xi\in Q,\quad \beta>0.
\]
The word transformation is $\rho_n(m)(a)=W_OF_m(q(a))$.
To see the size of this function family, fix $n$ distinct keys,
let $Q$ contain an open set, and allow the values to vary freely.
The scalar weight functions are linearly independent: they share
a positive denominator, and their numerators restrict on a
suitably chosen line to exponentials with distinct exponents.
Consequently, varying the values alone produces an
$np$-dimensional subspace of $C(Q,\mathbb R^p)$, whose elements
are generally nonlinear functions of the query. This calculation
concerns the head before output projection; the distinctions
visible on word states also depend on the reachable queries,
values, and $W_O$. The usual implementation stores $n(r+p)$
scalars and evaluates one query in $O(n(r+p))$ operations.
Evaluating all $n$ queries therefore has quadratic cost in
sequence length \cite{vaswani2017attention}.

\noindent\textbf{Linear attention: a response family of bounded
dimension.}
Let $\phi$ have $s$ positive feature coordinates. Normalized
linear attention represents the same ordered memory by
$z=\sum_j\phi(k_j)$ and $S=\sum_j\phi(k_j)v_j^\top$, giving
\[
F_m^{\mathrm{lin}}(\xi)
=
\frac{S^\top\phi(\xi)}{\varepsilon+z^\top\phi(\xi)}.
\]
The normalization means that this function need not be linear
in $\xi$. Its restriction comes from the factorization through
$(z,S)$. Indeed, for fixed keys put
$\Phi=[\phi(k_1)\mid\cdots\mid\phi(k_n)]$ and
$V=[v_1\mid\cdots\mid v_n]$. The denominator is fixed and the
values enter only through $S^\top=V\Phi^\top$. Varying the
values therefore produces a response family of dimension at
most $p\,\operatorname{rank}(\Phi)\leq ps$. When $n>s$, some
value variations distinguishable by the softmax head above
necessarily become invisible to this representation. In return,
the running state uses $s(p+1)$ scalars, and both appending a
record and evaluating a query cost $O(s(p+1))$ operations,
independently of $n$, once the features are computed
\cite{katharopoulos2020transformers}.

\noindent\textbf{Regression memory: spending computation on the
choice of response.}
Intention suggests another way to allocate these resources.
The memory determines a linear function fitted to the
associations $F(k_j)\approx v_j$. With
$G=\sum_j k_jk_j^\top$, $S=\sum_j k_jv_j^\top$, and
regularization $\gamma>0$, the fitted response is
$F_m^{\mathrm{reg}}(\xi)=S^\top(G+\gamma I)^{-1}\xi$
\cite{garnelo2023exploring}. The Gram matrix $G$ records how
the key directions overlap and changes how their associations
are read from $S$. The state size remains independent of $n$,
but the layer spends additional storage on $G$ and additional
computation on a linear solve. The gain comes from how memory
selects the transformation within the family of linear maps.

Mesa layers apply this regression principle to causal sequence
modeling. In matched-backbone language-model experiments,
MesaNet reported lower validation perplexity than recurrent
baselines including DeltaNet and Gated DeltaNet, while spending
additional inference operations on the regression solve
\cite{vonoswald2026mesanet}. This suggests two distinct routes
to improving an attention layer. When a representation identifies
memories that the task must distinguish, one can enlarge the
family of representable responses. When a compact function
family already contains suitable responses, one can instead
improve the rule that selects and evaluates them. The design
objective is to preserve the distinctions the task needs at an
acceptable cost in both storage and computation.

\subsection{Selectivity and gradient stability}

A memory must sometimes respond to a single relevant record in
the presence of many distractors. In softmax attention, suppose
that the relevant key has a logit advantage $\Delta$ over $b$
distractors with equal logits. Its weight is
$p_\star=(1+b e^{-\Delta})^{-1}$. Maintaining a fixed weight
$p_\star\in(0,1)$ therefore requires the logit advantage to grow
as $\log b$. At the other extreme, as $p_\star$ approaches one,
its derivative with respect to the relevant logit,
$p_\star(1-p_\star)$, approaches zero. These two effects concern
different regimes of the same readout. Veli\v{c}kovi\'{c}
et al.\ study the dispersion of softmax weights with increasing
input size \cite{pmlr-v267-velickovic25a}, while Zhai et al.\
observe an association between attention entropy collapse and
unstable training \cite{pmlr-v202-zhai23a}.

\noindent\textbf{Selective retrieval through regression.}
Intention suggests addressing this problem through the choice of
the memory function $\rho_n(m)$. Recall from
Section~\ref{intention} that, before the output projection, the
regularized regression readout takes the form
\[
F_m(\xi)
  = VK^\top(G+\gamma I)^{-1}\xi
  = \sum_{j=1}^{n}
    \bigl[k_j^\top(G+\gamma I)^{-1}\xi\bigr]v_j,
\qquad G=KK^\top,\quad \gamma>0.
\]
The coefficient of each value depends on the geometry of the
whole key collection through $G$. These coefficients can be
negative and their sum is unconstrained. Thus a query can
select a relevant direction through the fitted regression map
without requiring a softmax weight to approach one
\cite[Definition~2.7]{garnelo2023exploring}.

For example, let $e_1,e_2$ be orthonormal, take one relevant
record $(e_1,1)$, and add $b$ copies of $(e_2,0)$. The Gram
matrix is $G=e_1e_1^\top+b e_2e_2^\top$, and hence
$F_m(\xi)=e_1^\top\xi/(1+\gamma)$. At the query $\xi=e_1$,
both the response and its directional derivative along $e_1$
equal $1/(1+\gamma)$ for every $b$. A dot-product softmax head
at fixed scale assigns the relevant record a weight tending to
zero as $b$ grows. The geometry matters here: if the zero-valued
distractors instead have key $e_1$, the regression response also
decreases, to $1/(1+b+\gamma)$. The question for a learned
layer is therefore whether its key geometry separates the
directions that a task requires it to retrieve. The performance
reported for causal regression in MesaNet motivates testing this
hypothesis on learned representations \cite{vonoswald2026mesanet}.

\subsection{Convergence of the layer update and training stability}

Recall the full sequence update $T_N=f^{\times N}\circ L_N$,
including the pointwise feedforward map. When successive layers
share parameters, their action is the iteration
$h^{(0)}=s$, $h^{(\ell+1)}=T_N(h^{(\ell)})$. If $T_N$ is
continuous and the iterates converge to $h^\ast$, then
$T_N(h^\ast)=h^\ast$: applying the attention and feedforward
maps once more reproduces every word state. The residual
$\|T_N(h^{(\ell)})-h^{(\ell)}\|$ measures how much the
sequence still changes at depth $\ell$.

To use this principle during training, one can keep the input
sequence $s$ available at each iteration and write the update
as $T_{N,\theta}(h;s)$. This allows the equilibrium to depend
on $s$ even when the update is contractive in $h$. Suppose that
for each fixed $s$ the smooth update maps a nonempty closed
convex domain into itself and satisfies
$\|D_hT_{N,\theta}(h;s)\|\leq\kappa<1$ throughout that domain.
The iterates then converge to a unique equilibrium $h^\ast(s)$.
Writing $J_\ast=D_hT_{N,\theta}(h^\ast(s);s)$, implicit
differentiation gives
\[
D_\theta h^\ast(s)
 = (I-J_\ast)^{-1}
   \partial_\theta T_{N,\theta}(h^\ast(s);s),
\qquad
\|(I-J_\ast)^{-1}\|\leq\frac{1}{1-\kappa}.
\]
The same contraction bound therefore controls both convergence
of the forward iteration and amplification of parameter
perturbations through the equilibrium. Deep equilibrium models
use this fixed-point formulation for weight-tied sequence
layers \cite{bai2019deep}. Bai et al.\ subsequently showed that
regularizing the update Jacobian near equilibrium improved the
stability of both forward and backward solves
\cite{pmlr-v139-bai21b}. In our framework, this suggests applying
such regularization to the full update $T_N$, where it measures
the combined effect of memory formation, attention readout, and
feedforward composition on training gradients.

\section{Experiments on ImageNet-1K}\label{sec:geometric-attention-experiments}

We evaluate three choices in the construction of the memory representation
$\rho_n\colon M_n\to\operatorname{Hom}(A,A)$.
In CAFormer we change how a query pairs with the keys. In softmax ViT we change
the function contributed by each source token. In linear ViT we widen the
matrix used to represent the action of a memory on a query.
We compare eight ImageNet-1K classification runs \cite{russakovsky2015ilsvrc},
one per architecture, all with seed~42. The runs use $224\times224$ inputs,
an effective batch of 1,024, and a 300-epoch learning-rate schedule with
20 warmup epochs and a peak rate of $10^{-3}$. The two softmax ViT runs
reached their planned 60-epoch stopping point. The other six runs ended
early. We report validation accuracy at the end of each epoch and compare
each variant with its baseline over their shared observations. Differences
between runs describe a single training seed.

\subsection{Changing the geometry in CAFormer-S18}\label{subsec:caformer-experiment}

CAFormer-S18 uses convolution in its first two stages and dot-product attention
in its last two \cite{yu2024metaformerbaselines}. We replace the attention
mixer in the nine blocks of stage~3 and the three blocks of stage~4.
The three variants give different meanings to a match between a query and a key.

\textbf{Homogeneous CAFormer.}
Lift each query and key to a line through the origin in $\mathbb R^{33}$.
The line through $(x,1)$ has unit representative
$u(x)=(x,1)/\sqrt{1+\|x\|^2}$. The overlap of two lines is
$\langle u(q),u(k)\rangle^2$. A softmax over keys, with a fixed positive
score scale, gives greater weight to lines meeting at a smaller angle.

\textbf{Grassmannian CAFormer.}
Reshape the same 32 coordinates into $X\in\mathbb R^{8\times4}$ and take
the four-plane in $\mathbb R^{12}$ spanned by the columns of
$\left[\begin{smallmatrix}X\\I_4\end{smallmatrix}\right]$.
Writing $P_x$ for its orthogonal projector, the score is proportional to
$\operatorname{tr}(P_qP_k)$, up to a constant common to each softmax row.
This overlap sums the squared cosines of the four principal angles between
the planes. Both projector constructions preserve the baseline's learned
query, key, value, and output projection shapes and its 26.34 million parameters.

\textbf{Two-action CAFormer.}
Each head produces two ten-dimensional queries $q_i^{(1)},q_i^{(2)}$,
one shared ten-dimensional key $k_j$, and two 32-dimensional values
$v_j^{(1)},v_j^{(2)}$. Action $r$ forms its own distribution
$\alpha_{ij}^{(r)}=\operatorname{softmax}_j
\langle q_i^{(r)},k_j\rangle$. The head returns
$\sum_{r=1}^2\sum_j\alpha_{ij}^{(r)}v_j^{(r)}$.
One source token can therefore receive different weights for its two value
vectors. The revised projections give 26.23 million parameters.

\begin{table}[!htbp]
\centering\small
\caption{CAFormer-S18 comparisons at the last epoch shared by each variant
and ordinary CAFormer-S18. Accuracy is validation top-1 in percent.
The signed difference is variant minus baseline, in percentage points (pp).}
\label{tab:caformer-matched}
\begin{tabular}{@{}lrrrr@{}}
\toprule
Variant & Epoch & Variant (\%) & Baseline (\%) & Difference (pp)\\
\midrule
Grassmannian & 12 & 57.300 & 58.022 & $-0.722$\\
Two-action   & 23 & 67.690 & 67.940 & $-0.250$\\
Homogeneous  & 41 & 73.484 & 73.366 & $+0.118$\\
\bottomrule
\end{tabular}
\end{table}

\begin{figure}[!htbp]
\centering
\includegraphics[width=\linewidth]{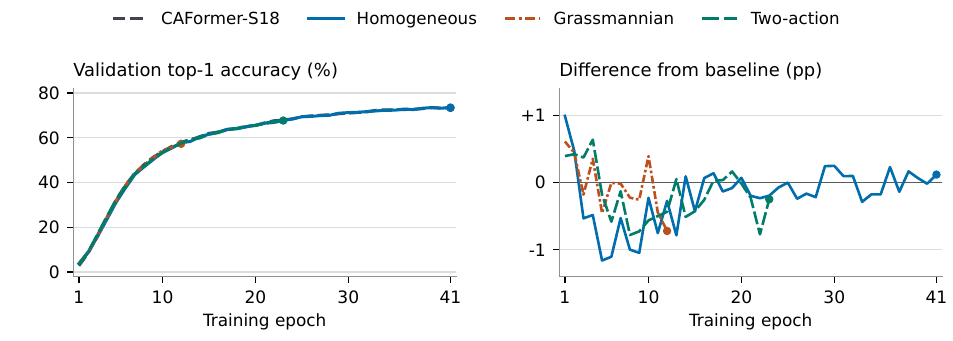}
\caption{CAFormer-S18 validation accuracy and its epoch-wise difference from
ordinary CAFormer-S18. Each variant ends at its last recorded epoch. The
baseline is shown through epoch~41 and continues alone to epoch~127.}
\label{fig:caformer-validation}
\end{figure}

At epoch~12, the last observation shared by all four runs, ordinary CAFormer
reaches 58.022\%, Homogeneous 57.748\%, Two-action 57.586\%, and Grassmannian
57.300\%. As the longer runs continue, the differences change sign
(Figure~\ref{fig:caformer-validation}). Homogeneous reaches 73.484\% at
epoch~41, within 0.12 points of the baseline. These results show that
attention based on line or subspace overlap can sustain accuracy close to
dot-product attention over the observed training intervals. The two-action
construction achieves comparable accuracy with two separate retrievals.

\subsection{Affine value functions in KV-action ViT}\label{subsec:kv-action-experiment}

In a softmax head, a source token contributes the same value $v_j$ wherever
it is read. Only its weight $\alpha_{ij}$ depends on the receiving query.
We give each token a second learned projection $c_j\in\mathbb R^{64}$ and
let its contribution vary with the query. For one head of ViT-5-Small/16
\cite{wang2026vit5}, the resulting memory acts by
\[
 F_j(q)=v_j+\operatorname{diag}(c_j)q,\qquad
 y_i=\sum_j\alpha_{ij}F_j(q_i)
     =\sum_j\alpha_{ij}v_j
       +q_i\odot\sum_j\alpha_{ij}c_j .
\]
Here $\odot$ is coordinatewise multiplication. The value $v_j$ is the
intercept, and $c_j$ supplies one slope for each query coordinate. The same
softmax weights aggregate both. The correction uses the projected query
$q_i$ before query normalization and rotary position encoding. The attention
scores in both models use the normalized, rotated queries and keys.

\begin{figure}[!htbp]
\centering
\includegraphics[width=\linewidth]{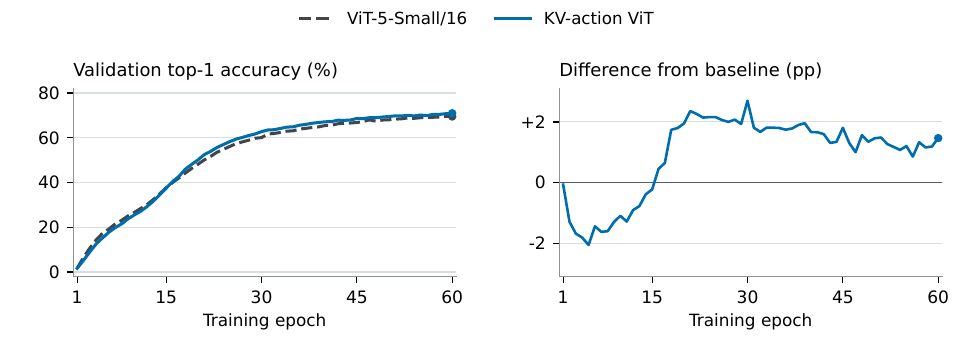}
\caption{KV-action ViT and ViT-5-Small/16 over all 60 recorded epochs. The
right panel shows when the affine-value model overtakes the baseline.}
\label{fig:kv-action-validation}
\end{figure}

The baseline leads through epoch~15. KV-action leads at every epoch from
16 through 60 (Figure~\ref{fig:kv-action-validation}). At epoch~60, KV-action
reaches 70.962\% top-1 versus 69.498\% for the baseline ($+1.464$ points).
Its top-5 accuracy is also higher (90.200\% versus 89.650\%), and its
validation loss is lower (1.268 versus 1.337).
This comparison supports using a query-dependent affine function to
represent a token's contribution. Two implementation details affect the
interpretation. The extra projection raises the parameter count from
22.04 to 23.81 million, and KV-action uses the unscaled dot product where
the baseline divides by $\sqrt{64}=8$. The measured gain therefore applies
to the affine-value architecture with this choice of score scale.

\subsection{Resizing the memory representation in linear ViT}\label{subsec:resized-linear-experiment}

For normalized linear attention, let $M_n$ consist of the ordered key--value
pairs for the $n$ tokens. At fixed token positions the head factors its
action through the statistics $(S,z)\in E_n$, where $E_n$ is the attainable
subset of $\mathbb R^{d_k\times64}\times\mathbb R^{d_k}$. Write $q_i,k_j$
for the normalized features before positional rotation and $R_i$ for the
rotation at position $i$. The implementation evaluates
\[
 S=\sum_{j=1}^n (R_jk_j)v_j^\top,\qquad
 z=\sum_{j=1}^n k_j,\qquad
 y_i=\frac{S^\top R_iq_i}{q_i^\top z+10^{-6}}.
\]
The numerator uses rotated features and the normalizer uses the positive
features before rotation. The output has 64 coordinates in both designs.

\textbf{Which memories can the readout distinguish?}
Let $\widetilde K$ have rows $(R_jk_j)^\top$ and let $V$ have rows $v_j^\top$.
At fixed keys, a change $\Delta V$ leaves the numerator matrix unchanged
exactly when $\widetilde K^\top\Delta V=0$. If the $n$ key rows are
independent, $S=\widetilde K^\top V$ determines every value at those keys.
The ordinary head allows at most 64 independent key rows. The resized head
raises this ceiling to 144 by constructing $k_j=a_j\otimes b_j$ from two
12-dimensional factors and projecting a full 144-dimensional query.
Rank-one tensors of this shape span $\mathbb R^{12}\otimes\mathbb R^{12}$.
The value $v_j=c_j\otimes d_j$ has factors of widths 16 and 4. This
factorization keeps the output width fixed while allocating more coordinates
to the dependence of the memory function on its query. At $224\times224$,
each head processes 201 tokens, so both constructions still compress the
full collection of token associations.

\begin{table}[!htbp]
\centering\small
\caption{Size of the linear-attention memory per image. The coordinate
counts include the normalizer $z$. Model parameters count learned weights
across the entire network.}
\label{tab:linear-memory-size}
\begin{tabular}{@{}lrr@{}}
\toprule
Quantity & Linear ViT & Resized linear ViT\\
\midrule
Shape of $S$ per head & $64\times64$ & $144\times64$\\
Coordinates of $z$ per head & 64 & 144\\
Coordinates of $(S,z)$ per head & 4,160 & 9,360\\
Coordinates across six heads per layer & 24,960 & 56,160\\
Trainable model parameters & 22,039,144 & 21,929,032\\
\bottomrule
\end{tabular}
\end{table}

\begin{figure}[!htbp]
\centering
\includegraphics[width=\linewidth]{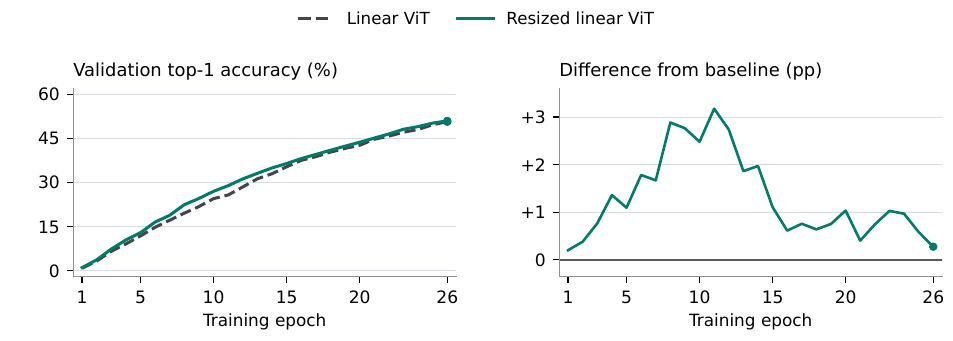}
\caption{Resized and ordinary linear ViT over the 26 epochs recorded for
both. The ordinary model continues alone through epoch~35.}
\label{fig:resized-linear-validation}
\end{figure}

The resized head uses $2.25$ times as many memory coordinates per image
with slightly fewer learned parameters (Table~\ref{tab:linear-memory-size}).
It leads in top-1 accuracy at all 26 matched epochs. Its largest lead is
3.176 points at epoch~11. By epoch~26 the gap narrows to 0.276 points
(50.994\% versus 50.718\%). The early accuracy advantage is consistent with
the greater number of key directions available at fixed value width.
The count measures the ambient space available to encode associations.
Both numerator matrices still have rank at most 64, and the resized model
also changes the key and value factorization and the positional rotations.
Determining how faithfully the trained $\rho_n$ distinguishes memories
would also require measuring which states the model reaches and which of
them induce distinguishable responses to queries.

\bibliographystyle{amsalpha}
\bibliography{references}
\end{document}